\documentclass{article}

\usepackage{microtype}
\usepackage{graphicx}
\usepackage{caption}
\usepackage{subcaption}
\usepackage{booktabs} 

\usepackage{algorithm}
\usepackage{algorithmic}

\usepackage{wrapfig}
\usepackage{amsfonts}       
\usepackage{nicefrac}       
\usepackage[round]{natbib}  
\usepackage{fullpage}
\usepackage{authblk}
\usepackage{hyperref}
\usepackage{url}            

\usepackage{enumitem}
\usepackage{multirow}
\usepackage{Definitions}
\usepackage{color}

\newcommand{\gp}{$\mathcal{GP}$}
\newcommand{\tp}{$\mathcal{TP}$}
\newcommand{\np}{$\mathcal{NP}$}
\newcommand{\ebp}{$\mathcal{EBP}$}
\newcommand{\vip}{$\mathcal{VIP}$}
\newcommand{\gpp}{$\mathcal{GPP}$}
\newcommand{\dpp}{$\mathcal{DPP}$}

\renewcommand{\geq}{\geqslant}
\renewcommand{\ge}{\geqslant}

\title{Energy-Based Processes for Exchangeable Data}

\author{
  $^*$Mengjiao Yang$^1$, \thanks{indicates equal contribution. Email: \texttt{\{sherryy, bodai\}@google.com}.}
  Bo Dai$^1$, Hanjun Dai$^1$, Dale Schuurmans$^{1,2}$\\ 
  
  $^1$Google Research, Brain Team, $^2$University of Alberta
}
\date{}
\begin{document}

\maketitle
\begin{abstract}
Recently there has been growing interest in modeling sets with exchangeability such as point clouds. A shortcoming of current approaches is that they restrict the cardinality of the sets considered or can only express limited forms of distribution over unobserved data. To overcome these limitations, we introduce \emph{Energy-Based Processes} (\ebp s), which extend energy based models to exchangeable data while allowing neural network parameterizations of the energy function. A key advantage of these models is the ability to express more flexible distributions over sets without restricting their cardinality. We develop an efficient training procedure for {\ebp s} that demonstrates state-of-the-art performance on a variety of tasks such as point cloud generation, classification, denoising, and image completion\footnote{The code is available at~{\href{https://github.com/google-research/google-research/tree/master/ebp}{https://github.com/google-research/google-research/tree/master/ebp}}.}.
\end{abstract}

\section{Introduction}\label{sec:intro}

Many machine learning problems consider data
where each instance is, itself, an unordered set of elements;
i.e., such that each observation is a \emph{set}.
Data of this kind
arises in a variety of applications,
ranging from document modeling \citep{BleNgJor03,garnelo2018conditional}
and
multi-task learning \citep{zaheer2017deep,edwards2016towards,liu2019neural}
to 3D point cloud modeling \citep{li2018point,yang2019pointflow}.
In unsupervised settings,
a dataset typically consists of a set of such sets,
while in supervised learning,
it consists of a set of (set, label) pairs.

Modeling a distribution over a space of instances,
where each instance is, itself, an unordered set of elements
involves two key considerations:
(1) the elements within a single instance are \emph{exchangeable},
i.e., the elements are order invariant;
and
(2) the cardinalities of the instances (sets) \emph{vary},
i.e., instances need not exhibit the same cardinality.
Modeling both \emph{unconditional} and \emph{conditional}
distributions over instances (sets) 
are relevant to consider,
since these support unsupervised and supervised tasks respectively.

For unconditional distribution modeling,
there has been significant prior work on modeling set distributions,
which has sought to balance competing needs to
expand model flexibility and preserve tractability on the one hand,
with respecting exchangeability and varying instance cardinalities
on the other hand.
However, managing these trade-offs has proved to be quite difficult,
and current approaches remain limited in different respects.

For example,
a particularly straightforward strategy for modeling distributions over
instances $\xb=\{x_1,...,x_n\}$ without assuming fixed cardinality
is simply to use a recurrent neural network (RNNs)
to encode instance probability auto-regressively via
$p\rbr{\xb}=\prod_{i=1}^np\rbr{x_i|x_{1:i-1}}$
for a permutation of its elements.
Such an approach allows the full flexibility of RNNs
to be applied, and has been empirically successful
\citep{larochelle2011neural,BahChoBen15},
but does not respect exchangeability
nor is it clear how to tractably enforce exchangeability with RNNs.

To explicitly ensure exchangeability, 
a natural idea has been
to
exploit De~Finetti's theorem,
which assures us that for any distribution over exchangeable elements
$\xb=\{x_1,...,x_n\}$
the instance probability can be decomposed as
\begin{eqnarray}\label{eq:definetti}
p\rbr{\xb}
& = & \int \prod_{i=1}^n p\rbr{x_i|\theta}p\rbr{\theta}d\theta
,
\footnotemark
\end{eqnarray}
\footnotetext{
De~Finetti's theorem is of course much more general than this,
establishing that any distribution over an \emph{infinitely} exchangeable 
sequence
can be equivalently expressed in the form \eqref{eq:definetti}.
}%
for some latent variable $\theta$.
In other words, 
there always exists a latent variable $\theta$ such that
conditioning on $\theta$ renders the instance elements $\cbr{x_i}_{i=1}^n$ \iid.
Latent variable models are therefore a natural choice
for expressing an exchangeable distribution.
Bayesian sets~\citep{GhaHel05}, 
latent Dirichlet allocation~\citep{BleNgJor03},
and related variants~\citep{BleLaf07,TehJorBeaBle06}
are classical examples of this kind of approach,
where the likelihood and prior in~\eqref{eq:definetti}
are expressed by simple known distributions. 
Although the restriction to simple distributions severely
limits the expressiveness of these models,
neural network parameterizations have recently been introduced 
\citep{edwards2016towards,KorDegHusGaletal18,yang2019pointflow}.
These approaches still exhibit limited expressiveness however:
\citet{edwards2016towards} 
restrict the model to \emph{known} distributions
parameterized by neural networks,
while
\citet{KorDegHusGaletal18,yang2019pointflow} 
only consider normalizing flow models
that require \emph{invertible} neural networks.

If we consider \emph{conditional} rather than \emph{unconditional} 
distributions over sets,
an extensive literature has considered stochastic process representations,
which exploits their natural exchangeability and consistency properties.
For example,
Gaussian processes~(\gp s)~\citep{RasWil06}
and extensions like Student-$t$ processes~(\tp s)~\citep{shah2014student},
are well known models that,
despite their scalability challenges,
afford significant modeling flexibility via kernels.
Unfortunately, they also
restrict the conditional likelihoods to simple known distributions.
\citet{DamLaw12,salimbeni2017doubly} enrich the expressiveness of {\gp s}
by stacking {\gp}-layers,
but at the cost of increasing inference intractability with increasing depth.
Neural processes~(\np s)~\citep{garnelo2018neural}
and subsequent variants~\citep{garnelo2018conditional,kim2019attentive}
attempt to construct neural network to mimic {\gp s},
but these too rely on \emph{known} distributions for the conditional likelihood,
which inherently limits expressiveness.

In this paper, we propose \emph{Energy-Based Processes}~(\ebp s),
and their extension to \emph{unconditional} distributions, 
to increase the flexibility of set distribution modeling
while retaining \emph{exchangeability} and \emph{varying-cardinality}.
After establishing necessary background on energy-based models (EBMs) and stochastic processes
in~\secref{sec:prelim},
we provide a new stochastic process representation theorem
in~\secref{sec:ebp}.
This result allows us to then generalize EBMs
to \emph{Energy-Based Processes}~(\ebp s),
which provably obtain the exchangeability and varying-cardinality properties.
Interestingly,
the stochastic process representation we introduce
also covers classical stochastic processes as simple special cases.
We further extend {\ebp} to the unconditional setting,
unifying the previously separate stochastic process and latent variable model
perspectives in a common framework.
To address the challenge of training {\ebp s},
we introduce an efficient new \emph{Neural Collapsed Inference}~(NCI)
procedure in~\secref{sec:deep_inference}.
Finally, we evaluate the effectiveness of {\ebp s} 
with NCI training on a set of \emph{supervised}
(\eg, $1$D regression and image completion) and 
\emph{unsupervised} tasks
(\eg, point-cloud feature extraction, generation and denoising),
demonstrating state-of-the-art performance
across a range of scenarios.

\section{Background}\label{sec:prelim}

We provide a brief introducton to energy-based models and stochastic processes,
which provide the essential building blocks for our subsequent development.

\subsection{Energy-Based Models}\label{subsec:ebm}

Energy-based models are attractive due to their flexibility
\citep{LecChoHadRanetal06,WuXieLuZhu18}
and appealing statistical properties~\citep{Brown86}.
In particular, an EBM over $\Omega\subset\RR^d$
with fixed dimension $d$ is defined as
\begin{eqnarray}\label{eq:ebm}
p_f\rbr{x} = \exp\rbr{f\rbr{x} - \log Z\rbr{f}}
\end{eqnarray}
for
$x\in\Omega$,
where
$f\rbr{x}:\Omega\rightarrow \RR$ is the energy function and
$Z\rbr{f}\defeq \int_\Omega \exp\rbr{f\rbr{x}}dx$ is the partition function.
We let $\Fcal\defeq \cbr{f\rbr{\cdot}: Z\rbr{f} < \infty}$.

The flexibility of EBMs is well known. 
For example, classical exponential family distributions 
can be recovered from 
\eqref{eq:ebm} by instantiating specific forms for $\Omega$ and
$f\rbr{\cdot}$.
Introducing additional structure to the energy function
allows both
Markov random fields~\citep{KinSne80}
and conditional random fields~\citep{LafMcCPer01} to 
be recovered from~\eqref{eq:ebm}.
More recently,
the introduction of deep neural energy functions
\citep{xie2016theory,DuMor19,dai2019exponential},
has led to many successful applications of EBMs 
to modeling complex distributions in practice.

Although maximum likelihood estimation estimation (MLE)
of general EBMs is notoriously difficult,
recent techniques such as adversarial dynamics embedding (ADE)
appear able to practically train a broader class of such models
\citep{dai2019exponential}.
In particular,
ADE approximates MLE for EBMs by formulating a saddle-point 
version of the problem:
\begin{equation}\label{eq:pd_ebm_mle}
	\max_{f}\min_{q\rbr{x, v}\in\Pcal}\,\, \widehat\EE\sbr{f\rbr{x}} - H\rbr{q(x, v)} - \EE_{q\rbr{x, v}}
\Big[f\rbr{x} - \frac{\lambda}{2}v^\top v\Big]
,
\end{equation}
where $p\rbr{x, v}$ is parametrized via a learnable Hamiltonian/Langevin
sampler.
Since we make use of some of the techniques in our main development,
we provide some further details of ADE in \appref{appendix:ade}.

Although these recent advances are promising,
EBMs remain fundamentally limited for our purposes,
in that they are only defined for fixed-dimensional data. 
The question of extending such models
to express distributions over
\emph{exchangeable} data with \emph{arbitrary cardinality}
has not yet been well explored.

\subsection{Stochastic Processes}\label{subsec:stoc_process}

Stochastic processes are usually  defined in terms of their finite-dimensional
marginal distributions.
In particular, consider a stochastic process given by a collection of random
variables $\cbr{X_{t}; t\in \Tcal}$ indexed by $t$, 
where the marginal distribution for any finite set of indices
$\cbr{t_1, \ldots, t_n}$ in $\Tcal$ (without order)
is specified
\ie, $p\rbr{x_{t_1:t_n}}\defeq p\rbr{x_{t_1},\ldots,x_{t_n}|\cbr{t_i}_{i=1}^n}$.
For example, Gaussian processes~(\gp s)
are defined in this way using Gaussians for the marginal distributions
\citep{RasWil06},
while Student-$t$ processes~(\tp s)
are similarly defined using
multivariate Student-$t$ distributions for the marginals
~\citep{shah2014student}.

The Kolmogrov extension theorem~\cite{oksendal2003stochastic}
provides the sufficient conditions for designing a valid stochastic processes,
namely:

\begin{itemize}
	\item {\bf Exchangeability} The marginal distribution for any finite set
of random variables is \emph{invariant to permutation} order.
Formally, for all $n$ and all permutations $\pi$, this means
	$$
	p\rbr{x_{t_1},\ldots, x_{t_n}|\cbr{t_i}_{i=1}^n} = p\rbr{\pi\rbr{x_{t_1:t_n}}|\pi\rbr{\cbr{t_i}_{i=1}^n}},
	$$
	where $p\rbr{\pi\rbr{x_{t_1:t_n}}}\defeq p\rbr{x_{\pi\rbr{t_1}},\ldots, x_{\pi\rbr{t_n}}}$. 
	\item {\bf Consistency} The partial mariginal distribution,
obtained by marginalizing additional variables in the finite sequence,
is the same as the one obtained from the original infinite sequence.
Formally, if $n\ge m\ge 1$, this means
	$$
	p\rbr{x_{t_1:t_m}|\cbr{t_i}_{i=1}^m} = \int p\rbr{x_{t_1:t_n}|\cbr{t_i}_{i=1}^n} dx_{t_{m+1}:t_n}. 
	$$
\end{itemize}

Obviously, these conditions also justify stochastic processes as a
valid tool for modeling exchangeable data.
However, existing classical models, such as \gp s and \tp s,
restrict the marginal distributions to simple forms 
while requiring huge memory and computational cost,
which prevents convenient application to
large-scale complex data.

\section{Energy-Based Processes}\label{sec:ebp}

We now develop our main modeling approach,
which combines a stochastic process representation 
of exchangeable data with energy-based models.
The result is a generalization of Gaussian processes and Student-t
processes that exploits EBMs for greater flexibility.
We follow this development with an extension to unconditional modeling.


\subsection{Representation of Stochastic Processes}\label{subsec:sp_rep}


Although finite marginal distributions provide a way to parametrize
stochastic processes,
it is not obvious how to use flexible EBMs to represent marginals
while still maintaining exchangeability and consistency.
Therefore, instead of such a direct parametrization,
we exploit the deeper structure of a stochastic process,
based on the following representation theorem.

\begin{theorem}\label{thm:representation}
For any stochastic process $\rbr{x_{t_1}, x_{t_2}, \ldots}\sim \Scal\Pcal$
that can be constructed via Kolmogrov extension theorem,
the process can be equivalently represented by a latent variable model
\begin{eqnarray}\label{eq:sp_lvm}
\theta\sim p\rbr{\theta},\quad 
x_{t_i}\sim p\rbr{x|\theta, t_i},\,\,\forall i\in\cbr{1,\ldots, n}
\;\forall n
,
\end{eqnarray}
where $\theta$ can be finite or infinite dimensional. 
\end{theorem}

Notice that $\theta$ can be either finite or infinite dimensional.
We use $p\rbr{\theta}$ to denote either the distribution or stochastic
process for the finite or infinite dimenstional random variable $\theta$
respectively. 

This is a straightforward corollary of De~Finetti's Theorem.
\begin{proof}
Since the process $\Scal\Pcal$ is constructed via
the Kolmogrov Extension Theorem,
it must satisfy exchangeability and consistency.
Therefore,
the sequence $\cbr{x_{t\rbr{1}},\ldots, x_{t\rbr{n}}}$ is exchangeable
$\forall n$. 
This implies,
by the conditional version of {De Finetti's Theorem},
that any marginal distribution can
be represented as a mixture of \iid processes: 
\begin{eqnarray}
p\rbr{x_{t_1:t_n}|\cbr{t_i}_{i=1}^n} = \int \prod_{i=1}^n p\rbr{x|\theta, {t_i}}p\rbr{\theta}d\theta.
\end{eqnarray}
Meanwhile, it is also easy to verify that such a model satisfies the
consistency condition. 
\end{proof}
Given such a representation of a stochastic processes,
it is now easy to see how to generalize Gaussian, Student-$t$,
and other processes with EBMs. 


\subsection{{\ebp} Construction}\label{subsec:ebp_construction}


To enhance the flexibility of a stochastic process representation
of exchangeable data,
we use EBMs to model the likelihood term in \eqref{eq:sp_lvm},
by letting
\begin{equation}
p_w\rbr{x|\theta, t}= \frac{\exp\rbr{f_w\rbr{x, t; \theta}}}{Z\rbr{f_w, t; \theta}}
,
\end{equation}
where $Z\rbr{f_w, t;\theta} = \int \exp\rbr{f_w\rbr{x, t;\theta}} dx$
and we let $w$ denote the parameters of $f$, which can be learned.
Substituting this into the latent variable representation of stochastic
processes~\eqref{eq:sp_lvm},
leads to the definition of energy-based processes on arbitrary finite
marginals as
\begin{equation}\label{eq:ebp}
\textstyle
p_w\rbr{x_{t_1:t_n}|\cbr{t_i}_{i=1}^n} = \int \frac{\exp\rbr{\sum_{i=1}^n\rbr{f_w\rbr{x_{t_i}, t_i; \theta}}}}{Z^n\rbr{f_w, t}}
p\rbr{\theta}d\theta
,
\end{equation}
given a prior $p\rbr{\theta}$ on the finite or infinite latent variable
$\theta$.
We refer to the resulting process as an energy-based process
(\ebp).

Compared to using restricted distributions, such as Gaussian or Student-$t$,
the use of an EBM in an \ebp\ allows much more flexible energy models $f_w$,
for example in the form of a deep neural network, to represent the complex dependency between $x$ and $t$. 
To rigoriously verify that the outcome is strictly more general than standard processes, observe that classical process models can be recovered exactly simply by
instantiating \eqref{eq:ebp} with specific choices of $f_w\rbr{x,t;\theta}$  and $p\rbr{\theta}$.

\begin{figure}[h]
	\centering
	\includegraphics[width=.7\linewidth]{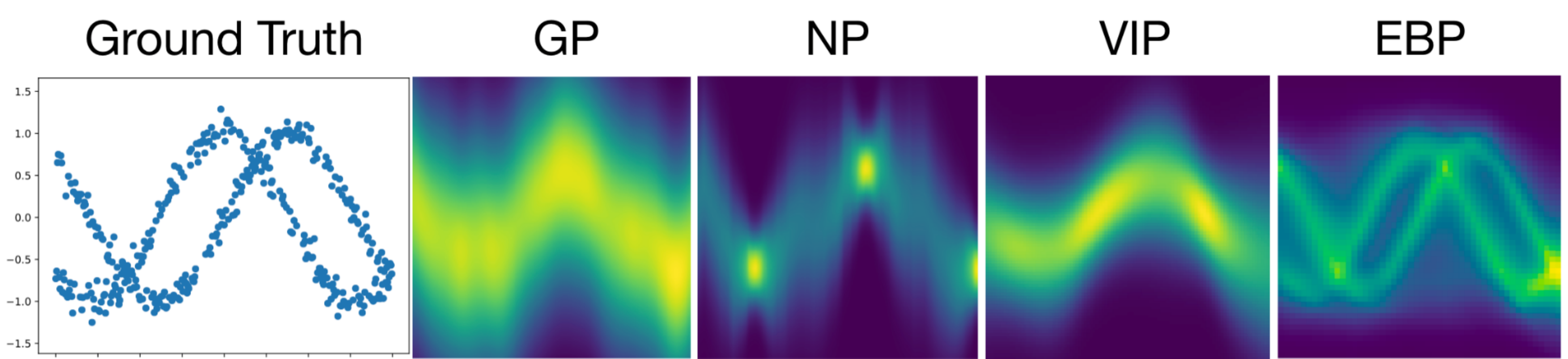}
	\caption{The ground truth data and learned energy functions of {\gp}, {\np}, {\vip}, and {\ebp} (from left to right). {\ebp} successfully captures multi-modality of the toy data as GP and NP exhibiting only a single mode;
see \secref{sec:experiment} for details.}
	\label{fig:exp_syn}
\end{figure}

\begin{itemize}
	\item {\bf Gaussian Processes}
Consider the weight-space view of {\gp s}~\citep{RasWil06},
which allows the {\gp} for regression to be re-written as
	\begin{eqnarray}\label{eq:gp_weight}
	{\theta} &\sim&  \Ncal\rbr{\zero, I_d},\\
	f_w\rbr{x, t; \theta} &=& -\frac{1}{2\sigma^2}\nbr{x - \theta^\top\phi\rbr{t}}^2,
	\end{eqnarray}
where $w = \cbr{\sigma, \phi\rbr{\cdot}}$,
with $\phi\rbr{\cdot}$ denoting feature mappings 
that can be finite or infinite dimensional. 
If we now let
$k\rbr{t, t'} = \phi\rbr{t}^\top \phi\rbr{t'}$
denote the kernel function and $K\rbr{t_{1:n}} = [k\rbr{t_i, t_j}]_{i,j}^n$,
the marginalized distribution can be recovered as
	$$
	p\rbr{x_{t_1:t_n}|\cbr{t_i}_{i=1}^n} = \Ncal\rbr{0, K\rbr{t_{1:n}} + \sigma^2 I_n},
	$$
which shows that
$X_t\sim \Gcal\Pcal\rbr{0, K\rbr{t_{1:n}} + \sigma^2 I_n}$;
see~\appref{appendix:lvm_gp}.

	\item {\bf Student-$t$ Processes}
Denote $\theta = \rbr{\alpha, \beta}$ and consider
	\begin{eqnarray}\label{eq:tp_weight}
	\textstyle
	\alpha&\sim& \Ncal\rbr{\zero, I_d},\,\, \beta^{-1} \sim \Gamma\rbr{\frac{\nu}{2}, \frac{\gamma}{2}},
\quad
\\
	f_w\rbr{x, t;\theta} &=& -\frac{\gamma\nbr{x - \sqrt{\frac{\beta\rbr{\nu - 2}}{\gamma}}\alpha^\top \phi\rbr{t}}^2}{2{\sigma^2\rbr{\nu - 2}\beta}},
	\end{eqnarray}
where $w = \cbr{\nu, \gamma,\sigma, \phi\rbr{\cdot}}$
with $\nu>0$ and $\gamma>0$.
These substitutions lead to the marginal distribution
	\begin{eqnarray*}
	p\rbr{x_{t_1:t_n}|\cbr{t_i}_{i=1}^n} = \Tcal\rbr{\nu, 0, K\rbr{t_{1:n}} + \sigma^2 I_n},
	\end{eqnarray*}
which shows that 
$X_t\sim\Tcal\Pcal\rbr{\nu, 0, , K\rbr{t_{1:n}} + \sigma^2 I_n}$;
see~\appref{appendix:lvm_tp}.

	\item {\bf Neural Processes}
Neural processes~(\np s)
are explicitly defined by a latent variable model in
\citep{garnelo2018neural}:%
\footnote{
The conditional neural processes~\citep{garnelo2018conditional}
only defines the predictive distribution, hence it is not a proper
stochastic processes, as discussed in their paper.
}
	\begin{eqnarray*}
	p\rbr{x_{t_1:t_n}|\cbr{t_i}_{i=1}^n} = \int \prod_{i=1}^n \Ncal\rbr{x|h_w\rbr{t_i; \theta}}p\rbr{\theta}d\theta,
	\end{eqnarray*}
where $h_w\rbr{\cdot; \theta}$ is a neural network.
Clearly,
\np s~share similarity to {\ebp s} in that both processes
use deep neural networks to enhance modeling flexibility.
However, there remain critical differences.
In fact, the likelihood function $p\rbr{x|t, \theta}$ in \np s
is still restricted to \emph{known} simple distributions,
with parameterization given by a neural network.
By contrast, {\ebp s} directly use EBMs with deep neural energy functions
to model the likelihood.
In this sense, {\ebp s} are a strict generalization of {\np s}:
if one fixes the last layer of $f_w$ in {\ebp s} to be a simple function,
such as quadratic,
then an {\ebp} reduces to a {\np}. 
\end{itemize}
\figref{fig:exp_syn}
demonstrates the comparison between these process models and an additional variational implicit process~(\vip) model (see~\appref{appendix:related_work})
in a simple regression setting,
highlighting the flexibility of {\ebp s}
in modeling the conditional likelihood.

\subsection{Unconditional {\ebp s} Extension}


Stochastic processes, such as {\ebp s}, express the \emph{conditional}
distribution over $\cbr{X_t}$ \emph{conditioned on} an index variable $t$,
which makes this approach naturally applicable to supervised learning tasks
on exchangeable data. 
However, we would also like to tackle unsupervised learning problems
given exchangeable observations,
so an \emph{unconditional} formulation of the \ebp\ is required.

To develop an unconditional \ebp, 
we start with the distribution of an arbitrary finite marginal,
$p\rbr{x_{t_1:t_n}|\cbr{t_i}_{i=1}^n}$.
Note that when the indices $\cbr{t_i}_{i=1}^n$ are not observed,
we can simply marginalize them out to obtain
\begin{align}\label{eq:ebp_unconditional}
p_w\rbr{x_{1:n}}&\defeq \int p_w\rbr{x_{t_1:t_n}|\cbr{t_i}_{i=1}^n} p\rbr{\cbr{t_i}_{i=1}^n} dt_{1:n}\quad\\
&=\int p_w\rbr{x_{t_1:t_n}|\cbr{t_i}_{i=1}^n, \theta}p\rbr{\theta}p\rbr{\cbr{t_i}_{i=1}^n}d\theta dt_{1:n}\nonumber
.
\end{align} 
Here we can introduce parameters to the $p\rbr{\cbr{t_i}_{i=1}^n}$,
which can also be learned.
It can be verified the resulting distribution
$p_w\rbr{x_{1:n}}$ is provably \emph{exchangeable} and \emph{consistent}
under mild conditions.

\begin{theorem}\label{thm:prior_consistent}
If $n\ge m\ge 1$, and the prior is exchangeable and consistent,
then the marginal distribution $p\rbr{x_{1:n}}$ will be exchangeable and consistent. 
\end{theorem}

The proof can be found in~\appref{appendix:proof_prior}. 

We refer to this result as the unconditional \ebp.
This understanding allows connections to be established with
some existing models.
\begin{itemize}

\item {\bf \gp-Latent Variable Model}
The \gp-latent variable model~(GPLVM)~\citep{Lawrence04}
considers the estimation of the latent index variables
by maximizing the $\log$-marginal likelihood of \gp, \ie, 
	\begin{equation}
		\max_{\cbr{t_i}_{i=1}^n}\,\, \log p\rbr{x_{t_1:t_n}|\cbr{t_i}_{i=1}^n}
		= \log \Ncal\rbr{\zero, K\rbr{t_{1:n}} + \sigma^2 I_n}.
	\end{equation}
This can be understood as using a point estimator with {\gp s}
and an improper uniform prior
$p\rbr{\cbr{t_i}_{i=1}^n}$ in~\eqref{eq:ebp_unconditional}.

\item {\bf Bayesian Recurrent Neural Model}
\citet{KorDegHusGaletal18} propose a model BRUNO for 
modeling exchangeable data.
This model actually uses degenerate kernels to eliminate
$\cbr{t_i}_{i=1}^n$ in~\eqref{eq:ebp_unconditional}.
In particular, BRUNO defines a {\tp} for each latent variable dimension,
with the same constant feature mapping $\phi\rbr{t} = 1$, $\forall t$.
That is, for the $d$-th dimension in $x$,
$\forall d\in\cbr{0, \ldots, D}$,  
	\begin{eqnarray}\label{eq:bruno}
	p\rbr{x^d_{t_{1}:t_n} | \cbr{t_i}_{i=1}^n} = p\rbr{x^d_{1:n}}\sim \Tcal\rbr{\nu_d, \mu_d, K_d},
	\end{eqnarray}
since the kernel is $K_d\rbr{t_{1:n}} = \one\one^\top + \rbr{\sigma_d}^2 I_n$.
The observations are then transformed via an invertible function,
\ie, $x' = \psi\rbr{x}$ with
$\det\rbr{\frac{\partial \psi\rbr{x}}{\partial x}}$ invertible.

\item{\bf Neural Statistician}
\citet{edwards2016towards} essentially generalize latent Dirichlet
allocation~\citep{BleNgJor03} with neural networks.
The model follows~\eqref{eq:ebp_unconditional} with a sophisticated
hierarchical prior. 
However, by comparison with {\ebp s},
the likelihood function used in neural statistician
is still restricted in known simple distributions.
Meanwhile, it follows vanilla amortized inference.
We will show how {\ebp s} can work with a more efficient inference
scheme in the next section. 
\end{itemize}

We provide more instantiations in~\appref{appendix:special_ebp} and the related work in~\appref{appendix:related_work}.

\section{Neural Collapsed Inference for Deep {\ebp s}}\label{sec:deep_inference}

By incorporating EBMs in the latent variable representation of a stochastic
process, we obtain a family of flexible models
that can capture complex structure in exchangeable data 
for both \emph{conditional} and \emph{unconditional} distributions.
We can exploit deep neural networks in parametrizing the energy function
as in~\citet{xie2016theory,DuMor19,dai2019exponential},
leading to \emph{deep \ebp s}.
However, this raises notorious difficulties in inference and
learning as a consequence of flexibility. 
Therefore, we develop an efficient \emph{Neural Collapsed Inference}~(NCI)
method for unconditional deep {\ebp s}.
(For the inference and learning of conditional {\ebp s},
please refer to~\appref{appendix:ebp_cond_inference}.)


\subsection{Neural Collapsed Reparametrization}\label{subsec:neural_collapsed}


We first carefully analyze the difficulties in inference and learning through the empirical $\log$-marginal distribution of the general {\ebp s} on given samples $\Dcal = \cbr{x^i_{1:n}}_{i=1}^N$:
\begin{eqnarray}\label{eq:ebp_logmargin}
\textstyle
\max_{w}\,\, \widehat\EE_{\Dcal}\sbr{\log p_w\rbr{x_{1:n}} },
\end{eqnarray}
where $ p_w\rbr{x_{1:n}}$ is defined in~\eqref{eq:ebp_unconditional}.

There are several integrations that are not tractable
in~\eqref{eq:ebp_logmargin}
given a general neural network parameterized $f_w\rbr{x, t; \theta}$:
\begin{itemize}

	\item[1.] The partition function $Z\rbr{f_w, t, \theta} = \int \exp\rbr{f\rbr{x, t; \theta}} dx$ is intractable in $p\rbr{x_{t_1:t_n}|\cbr{t_i}_{i=1}^n, \theta}$;

	\item[2.] The integration over $\theta$ will be intractable for $p\rbr{x_{t_1:t_n}|\cbr{t_i}_{i=1}^n}$;

	\item[3.] The integration over $\cbr{t_i}_{i=1}^n$ will be intractable for $p\rbr{x_{1:n}}$.

\end{itemize}
One can of course use vanilla amortized inference with the
neural network reparameterization
trick~\citep{KinWel13,rezende2014stochastic} for each intractable component,
as in~\citep{edwards2016towards},
but this leads to an optimization over the approximate posteriors
$q\rbr{x|t, \theta}$ and $q\rbr{\theta, \cbr{t_i}_{i=1}^n}$.
The latter distribution requires a complex neural network architecture
to capture the dependence in $\cbr{t_i}_{i=1}^n$,
which is usually a significant challenge.
Meanwhile, in most unsuperivsed learning tasks,
such as point cloud generation and denoising,
one is only interested in $x_{1:n}$,
while $\cbr{t_i}_{i=1}^n$ is not directly used.
Since inference over $\cbr{t_i}_{i=1}^n$ is only an intermediate step,
we develop the following
\emph{Neural Collapsed Inference}
strategy (NCI).

Collapsed inference and sampling strategies have previously been proposed 
for removing nuisance latent variables that can be tractably eliminated,
to reduce computational cost and accelerate inference
\citep{TehNewWel07,Porteousetal08}.
Due to the intractability of
$$
p_w\rbr{x_{1:n}|\theta} = \int p_w\rbr{x_{t_1:t_n}|\theta, \cbr{t_i}_{i=1}^n} p\rbr{\cbr{t_i}_{i=1}^n}dt_{1:n},
$$
standard collapsed inference cannot be applied.
However, since deep EBMs are very flexible,
$p_{w'}\rbr{x_{1:n}|\theta}$
can be directly reparameterized
with \emph{another} EBM:

\begin{equation}\label{eq:ebp_general}
p_{w'}\rbr{x_{1:n}|\theta} 
\propto {\exp\rbr{f_{w'}\rbr{x_{1:n}; \theta} } }
.
\end{equation}

Concretely, assume 
$p\rbr{\cbr{t_i}_{i=1}^n}\propto \exp\rbr{\sum_{i=1}^n h_v\rbr{t_i}}$,
so we have 
\begin{align*}
\textstyle
p_{w'}\rbr{x_{1:n}|\theta} &= \prod_{i=1}^n \int p_w\rbr{x_{t_i}|\theta, t_i}p\rbr{t_i} dt_i\\
&\propto \prod_{i=1}^n \int {\exp\rbr{f_w\rbr{x_{t_i}, t_i; \theta} - Z\rbr{f_w, t_i;\theta} + h_v\rbr{t_i}}} dt_i\\
&\approx \prod_{i=1}^n \frac{1}{Z\rbr{f_{w'}; \theta}}\exp\rbr{f_{w'}\rbr{x_i;\theta}},
\end{align*}
where the last step follows because the result of the integration in the
second step is a distribution $p\rbr{x}$ over $\Omega$,
and we are using another learnable EBM to approximate this distribution.
Therefore, we consider the collapsed model:
\begin{equation}\label{eq:ebp_rep}
p_{w'}\rbr{x_{1:n}|\theta} \propto {\exp\rbr{\sum_{i=1}^n f_{w'}\rbr{x_i; \theta} } },
\end{equation}
which still satisfies exchangeability and consistency.
In fact, with the \iid~prior on $\cbr{t_i}_{i=1}^n$,
we will obtain a latent variable model based on De Finetti's theorem.
With such an approximate collapsed model,
the $\log$-marginal distribution can be used as a surrogate:
\begin{eqnarray}\label{eq:ebp_collapsed_mle}
\ell\rbr{w}\defeq \log p_{w'}\rbr{x_{1:n}} = \log \int p_{w'}\rbr{x_{1:n}|\theta}p\rbr{\theta} d\theta.
\end{eqnarray}

We refer to the variational inference in such a task-oriented
neural reparametrization model as \emph{Neural Collapsed Inference},
which reduces the computational cost and memory of inferring the posterior
compared to using vanilla variational amortized inference.

We can further use the neural collapsing trick for $\theta$; which will reduce the model to Gibbs point processes~(\gpp s)~\citep{Dereudre19} and Determinantal point processes~(\dpp s)~\citep{lavancier2015determinantal,kulesza2012determinantal}. Therefore, the proposed algorithm can straightforwardly applied for deep~\gpp~and~\dpp~estimation. It should be emphasized that by exploiting the proposed primal-dual MLE framework, we automatically obtain a deep neural network parametrized dual sampler with the learned model simultaneously, which can be used in inference and bypass the notorious sampling difficulty in \gpp~and \dpp. Please see~\appref{appendix:ebp_further_inference} for detailed discussion.


\subsection{Amortized Inference}\label{subsec:pd_inference}


As discussed, $\cbr{t_i}_{i=1}^n$ can be eliminated by
neural collapsed reparameterization.
We now discuss variational techniques for integrating over
$\theta$ and $x$ respectively in the partition function
of~\eqref{eq:ebp_collapsed_mle}

\noindent{\bf ELBO for integration on $\theta$}
We apply vanilla ELBO to handle the intractability of integration over $\theta$.
Specifically, since
\begin{eqnarray}\label{eq:elbo}
\log \int p_{w'}\rbr{x_{1:n}|\theta}p\rbr{\theta} d\theta = \max_{q\rbr{\theta|x_{1:n}}\in\Pcal} \EE_{q\rbr{\theta|x_{1:n}}}\sbr{\log p_{w'}\rbr{x_{1:n}|\theta}} - KL\rbr{q||p}
,
\end{eqnarray}
we can apply the standard reparameterization trick~\citep{KinWel13,rezende2014stochastic} for $q\rbr{\theta|x_{1:n}}$.

\noindent{\bf Primal-Dual form for partition function}
For the term $\log p_{w'}\rbr{x_{1:n}|\theta}$ in~\eqref{eq:elbo}, which is
$$
\log p_{w'}\rbr{x_{1:n}|\theta} = f_{w'}\rbr{x_{1:n}; \theta} - \log Z\rbr{f_{w'}, \theta},
$$
we apply an adversarial dynamics embedding
technique~\citep{dai2019exponential} for the $\log Z\rbr{f_{w'}, \theta}$
as introduced in~\secref{sec:prelim}.
This leads to an equivalent optimization of the form 
\begin{equation}\label{eq:pd_logz}
\log  p_{w'}\rbr{x_{1:n}|\theta}
\propto \min_{q\rbr{x_{1:n}, v|\theta}\in\Pcal} f_{w'}\rbr{x_{1:n}; \theta} - H\rbr{q\rbr{x_{1:n}, v|\theta}}
	- \EE_{q\rbr{x_{1:n}, v|\theta}}\sbr{f_{w'}\rbr{x_{1:n};\theta} - \frac{\lambda}{2}v^\top v}.
\end{equation}
By combining \eqref{eq:elbo} and~\eqref{eq:pd_logz}
into~\eqref{eq:ebp_collapsed_mle},
we obtain
\begin{equation}\label{eq:ebp_obj}
\max_{w', q\rbr{\theta|x_{1:n}}}\min_{q\rbr{x_{1:n}, v|\theta}} L\rbr{q\rbr{\theta|x_{1:n}}, q\rbr{x_{1:n}, v|\theta}; w'},
\end{equation}
where 
\begin{align*}
L\rbr{q\rbr{\theta|x_{1:n}}, q\rbr{x_{1:n}, v|\theta}; w'}
& \defeq \widehat\EE_{x_{1:n}}\EE_{q\rbr{\theta|x_{1:n}}}\sbr{f_{w'}\rbr{x_{1:n};\theta}}\\
& - \widehat\EE_{x_{1:n}}\EE_{q\rbr{\theta|x_{1:n}}}\sbr{\EE_{q\rbr{x_{1:n}, v|\theta}}\sbr{f_{w'}\rbr{x_{1:n};\theta} - \frac{\lambda}{2}v^\top v}}\\
& -\widehat\EE_{x_{1:n}}\sbr{H\rbr{q\rbr{x_{1:n}, v|\theta}} - KL\rbr{q\rbr{\theta|x_{1:n}}||p\rbr{\theta}}}
.
\end{align*}

\begin{algorithm}[t] 
\caption{Neural Collapsed Inference} \label{alg:ebp_mle}
  \begin{algorithmic}[1]
    \STATE Initialize $W_1$ randomly, set length of steps $T$. 
    \FOR{iteration $k=1, \ldots, K$}
        \STATE Sample mini-batch ${\cbr{x^j_{1:n_j}}_{j=1}^b}$ from dataset $\Dcal$.
        \STATE Sample $\theta^j\sim q_\alpha\rbr{\theta|x_{1:n}}$, $\forall j=1,\ldots,b$. 
        \STATE Sample ${\xtil^j_{1:n}, \vtil^j}\sim q_\beta\rbr{x_{1:n}, v|\theta}$, $\forall j=1,\ldots,b$. 
      \STATE $\cbr{\beta_{k+1}} =\beta_k-\gamma_k\hat{\nabla}_{\beta}L\rbr{\alpha_k, \beta_k, w'_k}$
      \STATE $\cbr{\alpha, w'}_{k+1}=\cbr{\alpha, w'}_k+\gamma_k \hat{\nabla}_{\cbr{\alpha, w'}} L\rbr{\alpha_k, \beta_k; w'_k}$. 
    \ENDFOR
  \end{algorithmic}
\end{algorithm}
\noindent{\bf Parametrization}
Finally, we describe some concrete parameterizations for
$f_w\rbr{x;\theta}$, $q\rbr{\theta|x_{1:n}}$ and $ q\rbr{x_{1:n}, v|\theta}$. 

The energy function $f_w\rbr{x;\theta}$ is parametrized as a MLP
that takes input $x_{t_i}$ concatenated with $\theta$.
We use the same energy function parameterization
for both conditional and unconditional {\ebp s}.  

For $q\rbr{\theta|x_{1:n}}$ we use a simple Gaussian with
mean function parameterized via deepsets~\citep{zaheer2017deep}:
\begin{eqnarray}\label{eq:qtheta_param}
\theta = \mathtt{mlp}_{\alpha}\rbr{x_{1:n}} + \sigma\xi,\,\, \xi\sim \Ncal\rbr{0, I_d}
,
\end{eqnarray}
where $\mathtt{mlp}_{\alpha}\rbr{x_{1:n}}\defeq \sum_{i=1}^n\phi\rbr{x_i}$
and $\alpha$ denoting the parameters in $\phi\rbr{\cdot}$.

For $q\rbr{x_{1:n}, v|\theta}$ we consider dynamics embedding with 
an {RNN} or flow-model as the initial distribution;
see~\appref{appendix:ade_param} for parameterization
and~\appref{appendix:exp_details} for implementation details.
We denote the parameters in $q\rbr{x_{1:n}, v|\theta}$ as $\beta$.
We also denote the objective in~\eqref{eq:ebp_obj} as
$L\rbr{\alpha, \beta; w'}$.
Then, we can use stochastic gradient descent for~\eqref{eq:ebp_obj}
to optimize
$W = \rbr{\alpha, \beta, w'}$, as illustrated in~\algref{alg:ebp_mle}.

\section{Applications}\label{sec:experiment}
We test conditional {\ebp s} on two supervised learning tasks: 1D regression and image completion, and unconditional {\ebp s} on three unsupervised tasks: point cloud generation, representation learning, and denoising. Details of each experiment can be found in~\appref{appendix:exp_details}.

\subsection{Supervised Tasks}
\paragraph{1D regression.}
In order to show that {\ebp s} are more flexible than {\gp s}, {\np s} and {\vip s} in modeling complex distributions, we construct a two-mode synthetic dataset of $i.i.d.$ points whose means form two sine waves with a phase offset. In this setting, $t_i$ corresponds to the horizontal axis of the sine wave and $x_{t_i}$ corresponds to the values on the vertical axis. At every training step, we randomly select a subset of the points as observations and estimate the marginal distribution of the observed and unobserved points similar to~\citet{garnelo2018neural}. We visualize the ground truth and learned energy functions of {\gp}, {\np}, {\vip} and {\ebp} in~\figref{fig:exp_syn}. Clearly, {\ebp} succeeds as {\gp} and {\np} fail to capture the multi-modality of the underlying data distribution. More comparisons can be found in~\appref{sec:appx_syn}.

\paragraph{Image completion.}
An image can be represented as a set of $n$ pixels $\cbr{(t_i, x_{t_i})}_{i=1}^n$, where $t_i \in \RR^2$ corresponds to the Cartesian coordinates of each pixel and $x_{t_i}$ corresponds to the channel-wise intensity of that pixel ($x_{t_i} \in \RR$ for grayscale images and $x_{t_i} \in \RR^3$ for RGB images). Conditional {\ebp s} perform image completion by maximizing $p(x_{t_i:t_n} | \cbr{t_i}_{i=1}^n)$.

\begin{figure}[t]
\centering
  \includegraphics[width=.8\columnwidth]{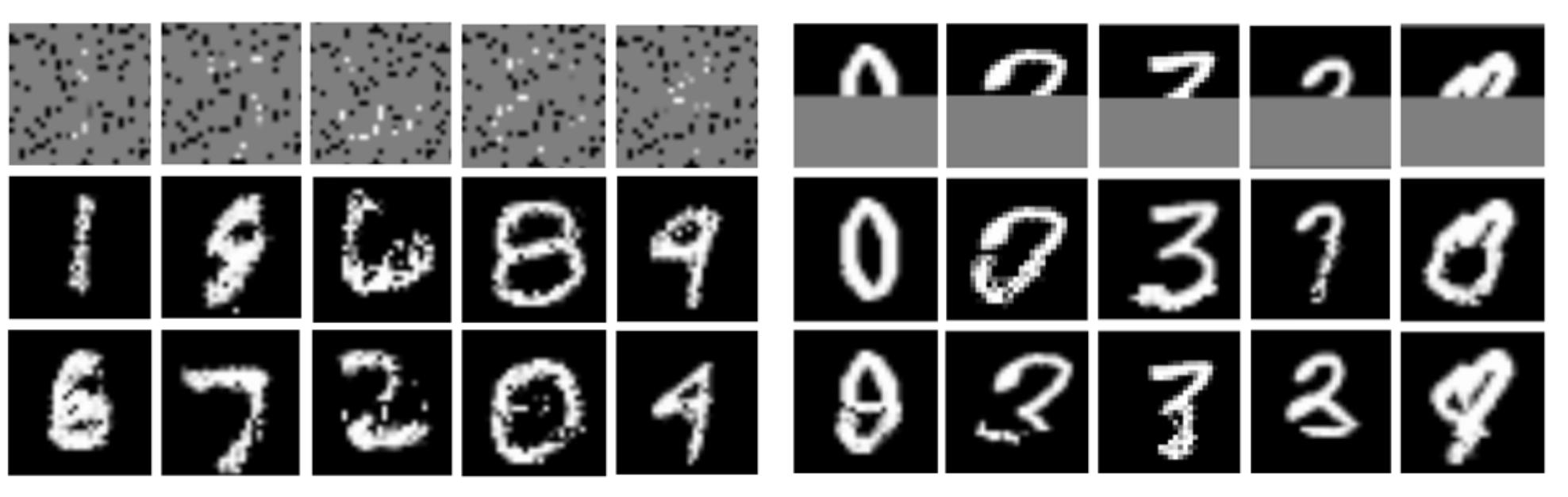}
  \caption{Image completion on MNIST. The first row shows the unobserved pixels in gray and observed pixels in black and white. The second and third rows are two different generated samples given the observed pixels from the first row. Generations are based on randomly selected pixels or the top half of an image.}
  \label{fig:exp_comp_mnist}
\end{figure}

\begin{figure}
  \centering
  \includegraphics[width=.8\columnwidth]{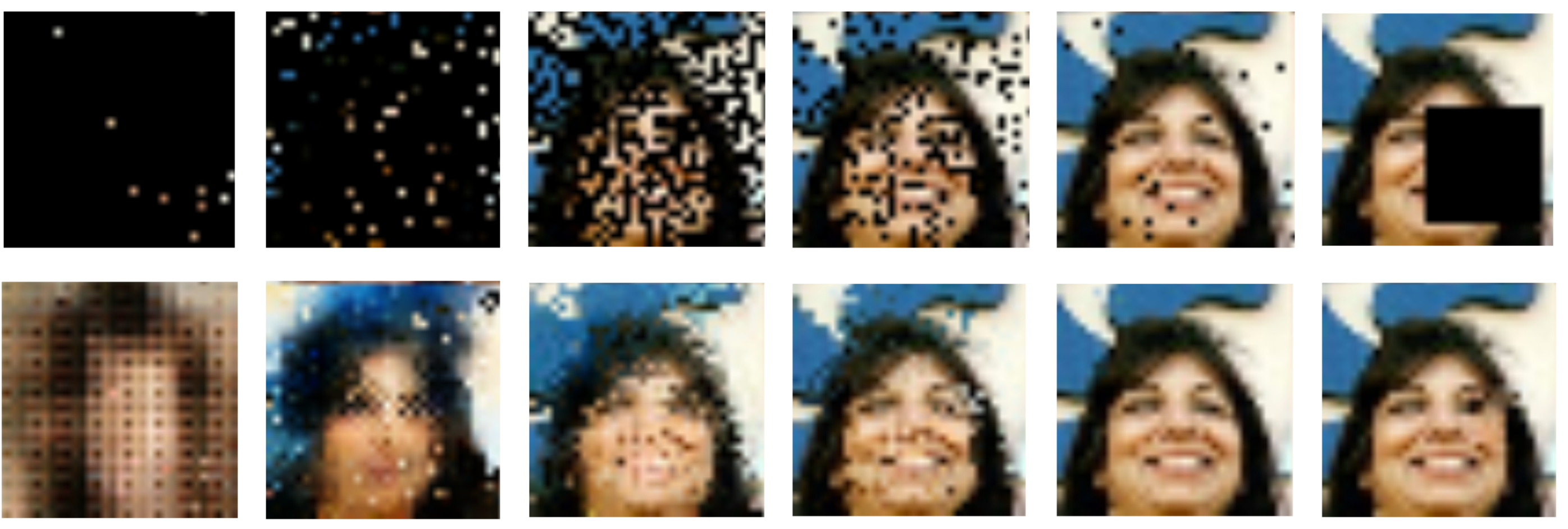}
  \caption{Image completion on CelebA. The first row shows the unobserved pixels in black with an increasing number of observed pixels from left to right (column 1-5). The second row shows the completed image given the observed pixels from the first row.}
  \label{fig:exp_comp_celeba}
\end{figure}

We separately train two conditional {\ebp s} on the MNIST~\citep{LeCun98} and the CelebA dataset~\citep{liu2015deep}. Examples of completion results are shown in~\figref{fig:exp_comp_mnist} and~\figref{fig:exp_comp_celeba}. When a random or consecutive subset of pixels are observed, our method discovers different data modes and generates different MNIST digits, as shown in~\figref{fig:exp_comp_mnist}. When a varying number of pixels are observed as in~\figref{fig:exp_comp_celeba}, completion with fewer observed pixels (column 2) can lead to a face that is much different from the original face than completion with more observed pixels (column 5), revealing high variance when the number of observations is small (similar to \gp s). More examples of image completion can be found in~\appref{appendix:more_exp_results}.

\subsection{Unsupervised Tasks on Point Clouds}

Next, we apply unconditional {\ebp s} to a set of unsupervised learning tasks for point clouds. A point cloud represents a 3D object as the Cartesian coordinates of the set of exchangeable points $\cbr{x_i}_{i=1}^n \subset \RR^3$, where $n$ is the number of points in a point cloud and can therefore be arbitrarily large. Since the point cloud data does not depend on index $t_i$, they are modeled by unconditional {\ebp s} which integrate over $\cbr{t_i}_{i=1}^n$, leading to the unconditional objective $p\rbr{x_{1:n}} = \int p\rbr{x_{t_1:t_n}|\cbr{t_i}_{i=1}^n} p\rbr{\cbr{t_i}_{i=1}^n} d\cbr{t_i}_{i=1}^n$ as first introduced in~\eqref{eq:ebp_unconditional}.

\paragraph{Related work.}
Earlier work on point cloud generation and representation learning simply treats point clouds as matrices with a fixed dimension~\citep{achlioptas2017learning,gadelha2018multiresolution,zamorski2018adversarial,sun2018pointgrow}, leading to suboptimal parameterizations as permutation invariance and arbitrary cardrinality of exchangeable data are violated by this representation. Some of the more recent work tries to overcome the cardinality constraint by trading off flexibility of the model. For instance, ~\citet{yang2019pointflow} uses normalizing flow to transform an arbitrary number of points sampled from the initial distribution, but requires the transformations to be invertible. ~\citet{yang2018foldingnet}, as another example, transforms 2D distributions to 3D targets, but assumes that the topology of the generated shape is genus-zero or of a disk topology. 
~\citet{li2018point} demonstrate the straightforward extension of GAN is not valid for exchangeable data, and then, provide some strategies to make up such deficiency. However, the generator in the proposed PC-GAN is \emph{conditional} on observations, which restricts the usages of the model. 
Among all generative models for point clouds considered here, {\ebp s} are the most flexible in handling permutation-invariant data with arbitrary cardinality.

\paragraph{Point cloud generation.}
\begin{figure}
	\centering
	\includegraphics[width=.6\linewidth]{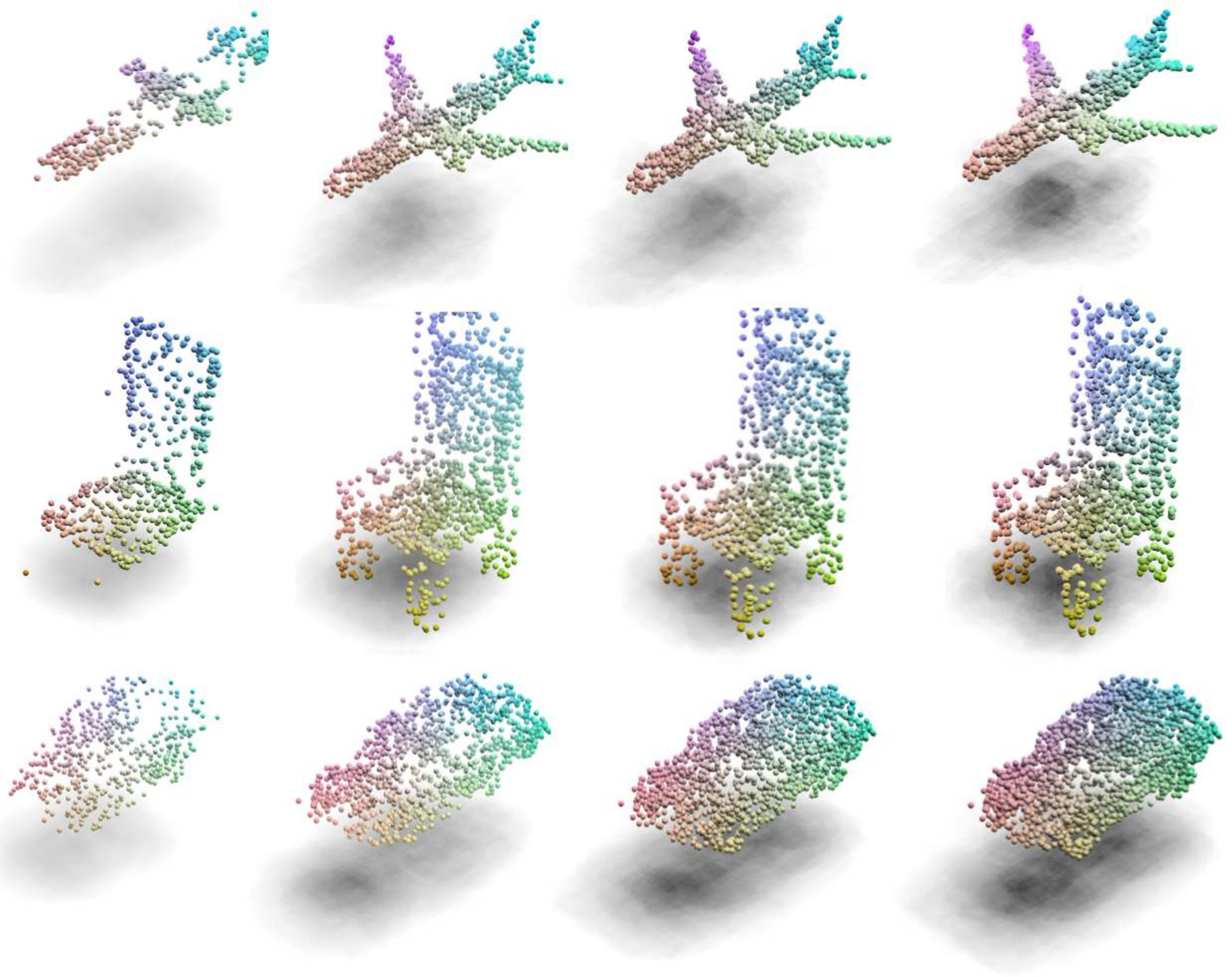}
	\caption{Example point clouds of airplane, chair, and car generated 
  from the learned model.}
\label{fig:exp_generation}
\end{figure}
\begin{figure}
	\centering
	\includegraphics[width=.6\linewidth]{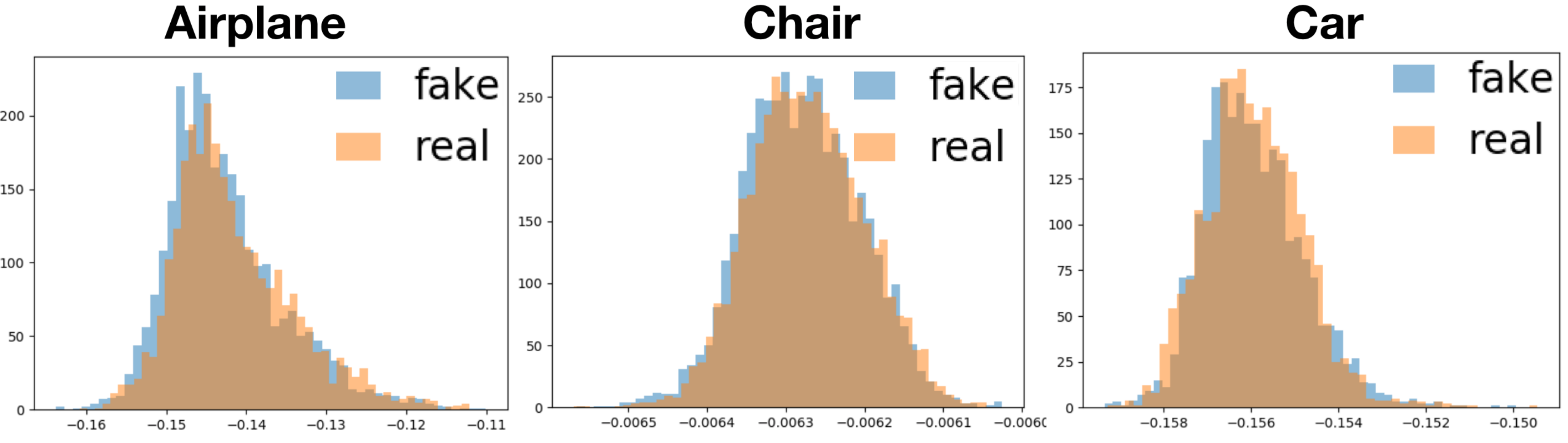}
	\caption{Energy distributions of the generated samples (fake) and training data (real). x-axis is the energy value and y axis is the count of examples. The energy distributions of the generated and real point clouds show significant overlap.}
	\label{fig:exp_energy}
\end{figure}

We train one unconditional {\ebp} per category on airplane, chair, and car from the ShapeNet dataset~\citep{WuSonKhoYuetal15}.~\figref{fig:exp_generation} shows the accumulative output of the model (see more generated examples in~\appref{appendix:more_exp_results}). We plot the energy distributions of all real and generated samples for each object category in~\figref{fig:exp_energy}. {\ebp s} have successfully learned the desired distributions as the energies of real and generated point clouds show significant overlap.

We compare the generation quality of {\ebp s} with the previous state-of-the-art generative models for point clouds including l-GAN~\citep{achlioptas2017learning}, PC-GAN~\citep{li2018point}, and PointFlow~\citep{yang2019pointflow}. Following these prior work, we uniformly sample 2048 points per point cloud from the mesh surface of ShapeNet, use both Chamfer distance (CD) and earth mover's distance (EMD) to measure similarity between point clouds, and use Jensen-Shannon Divergence (JSD), Minimum matching distance (MMD), and Coverage (COV) as evaluation metrics.~\tabref{table:generation} shows that {\ebp} achieves the best COV for all three categories under both CD and EMD, demonstrating {\ebp s} advantage in expressing complex distributions and avoiding mode collapse. {\ebp} also achieves the lowest JSD for two out of three categories. More examples of point cloud generation can be found in~\appref{appendix:more_exp_results}.

\begin{table}[h!]
  \setlength{\tabcolsep}{3pt}
    \caption{Generation results. $\uparrow$: the higher the better. $\downarrow$: the lower the better. The best scores are highlighted in bold. JSD is scaled by $10^2$, MMD-CD by $10^3$, and MMD-EMD by $10^2$. Each number for l-GAN is from the model trained using either CD or EMD loss, whichever one is better.}
\label{table:generation}

	\centering
	\begin{small}
    \begin{tabular}{llccccc}
    \toprule      
      & & \multirow{2}{*}{JSD ($\downarrow$)} & \multicolumn{2}{c}{ MMD ($\downarrow$) } & \multicolumn{2}{c}{ COV ($\%, \uparrow$) } \\
      \cmidrule(l){4-5}\cmidrule(l){6-7}
      Category & Model & & CD & EMD & CD & EMD \\
      \midrule
      \multirow{4}{*}{ Airplane } & l-GAN & \textbf{3.61} & 0.239 & 3.29 & 47.90 & 50.62 \\
      & PC-GAN & 4.63 & 0.287 & 3.57 & 36.46 & 40.94 \\
      & PointFlow & 4.92 & \textbf{0.217} & 3.24 & 46.91 & 48.40 \\
      & EBP (ours) & 3.92 & 0.240 & \textbf{3.22} & \textbf{49.38} & \textbf{51.60} \\
      \midrule
      \multirow{4}{*}{ Chair } & l-GAN & 2.27 & 2.46 & \textbf{7.85} & 41.39 & 41.69 \\
      & PC-GAN & 3.90 & 2.75 & 8.20 & 36.50 & 38.98 \\
      & PointFlow & 1.74 & \textbf{2.42} & 7.87 & 46.83 & 46.98 \\
      & EBP (ours) & \textbf{1.53} & 2.59 & 7.92 & \textbf{47.73} & \textbf{49.84} \\
      \midrule
      \multirow{4}{*}{ Car } & l-GAN & 2.21 & 1.48 & 5.43 & 39.20 & 39.77 \\
      & PC-GAN & 5.85 & 1.12 & 5.83 & 23.56 & 30.29 \\
      & PointFlow & 0.87 & \textbf{0.91} & \textbf{5.22} & 44.03 & 46.59 \\
      & EBP (ours) & \textbf{0.78} & 0.95 & 5.24 & \textbf{51.99} & \textbf{51.70} \\
      \bottomrule
    \end{tabular}
  \end{small}
\end{table}

\paragraph{Unsupervised representation learning.} Next, we evaluate the representation learning ability of {\ebp s}. Following the convention of previous work, we first train one {\ebp} on all 55 object categories of ShapeNet. We then extract the Deep Sets output ($\theta$ in our model) for each point cloud in ModelNet40~\citep{WuSonKhoYuetal15} using the pre-trained model, and train a linear SVM using the extracted features.~\tabref{table:classification} shows that our method achieves the second highest classification accuracy among the seven state-of-the-art unsupervised representation learning methods, and is only $0.1\%$ lower in accuracy than the best performing method. Since categories in ShapeNet and ModelNet40 only partially overlap, the representation learning ability of {\ebp s} can generalize to unseen categories.
\begin{table}[h]
  \centering
  \caption{Classification accuracy on ModelNet40. Models are pre-trained on ShapeNet before extracting features on ModelNet40. Linear SVMs are then trained using the learned representations.}  
  \label{table:classification}
    \begin{tabular}{cc}
      \toprule
      Model & Accuracy \\
      \midrule
      VConv-DAE~\citep{sharma2016vconv} & 75.5\\
      3D-GAN~\citep{wu2016learning} & 83.3 \\            
      l-GAN (EMD)~\citep{achlioptas2017learning} & 84.0 \\      
      l-GAN (CD)~\citep{achlioptas2017learning} & 84.5 \\
      PointGrow~\citep{sun2018pointgrow} & 85.7 \\
      MRTNet-VAE~\citep{gadelha2018multiresolution} & 86.4 \\
      PointFlow~\citep{yang2019pointflow} & 86.8 \\
      PC-GAN~\citep{li2018point} & 87.8 \\
      FoldingNet~\citep{yang2018foldingnet} & \textbf{88.4} \\
      \midrule
      EBP (ours) & 88.3 \\
      \bottomrule
\end{tabular}   
\end{table}

\begin{figure}[h!]
	\centering
	\includegraphics[width=.6\columnwidth]{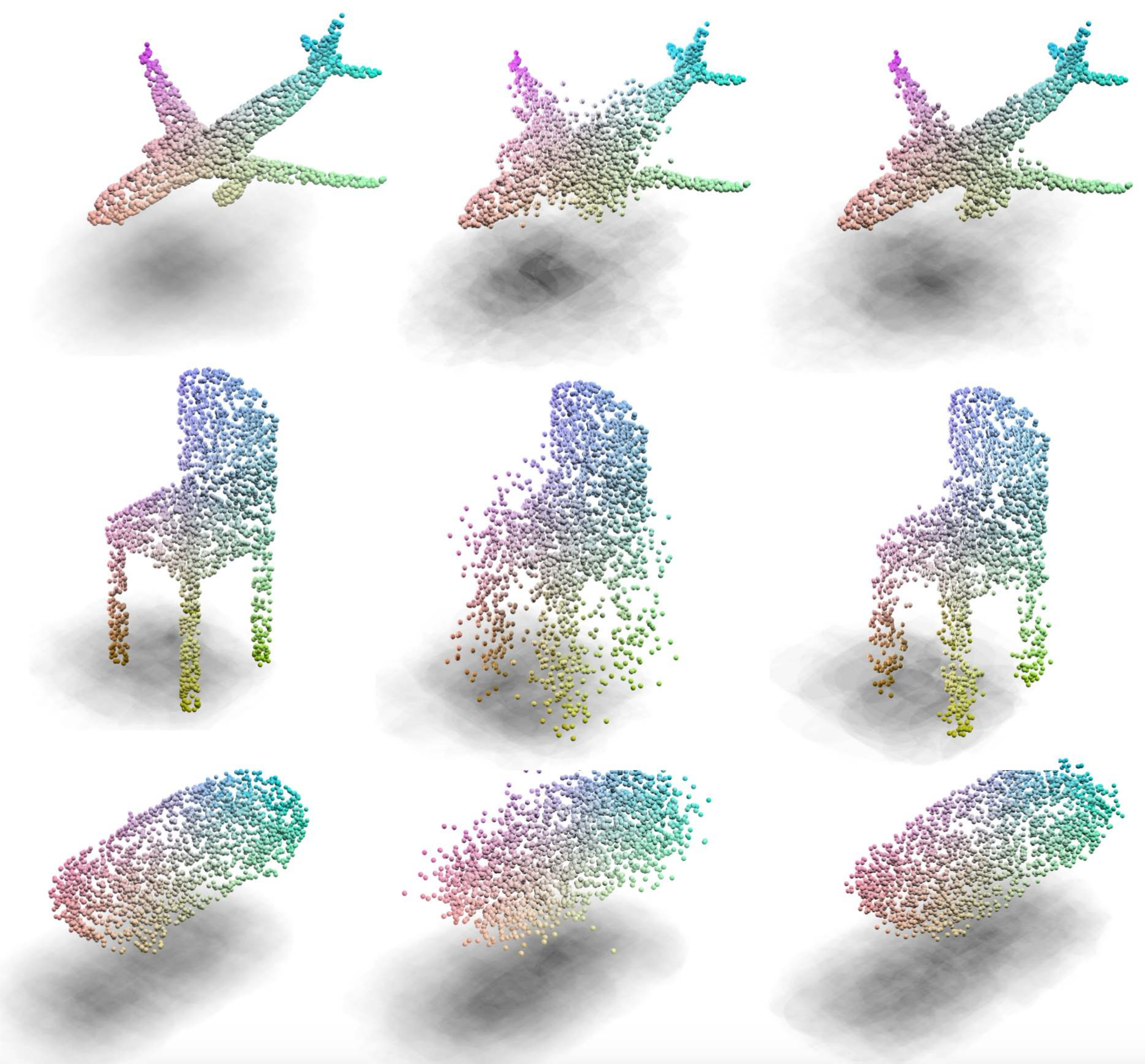}
	\caption{Examples of point cloud denoising using MCMC sampling. From left to right: original, perturbed, and denoised point clouds. }
	\label{fig:denoising}
\end{figure}

\paragraph{Point cloud denoising.}
Lastly, we apply {\ebp s} to point cloud denoising by running MCMC sampling using noisy point clouds as initial samples. To create noisy point clouds, we perturb samples from the initial distribution by selecting a random point from the set and add Gaussian perturbations to points within a small radius $r$ of the selected point. We then perform 20 steps of Langevin dynamics with a fixed step size while keeping the unperturbed points fixed. Results in ~\figref{fig:denoising} show that the gradient of our learned energy function is capable of guiding the MCMC sampling to recover the original point clouds. More examples of denoising can be found in~\appref{appendix:more_exp_results}.

\section{Conclusion}\label{sec:conclusion}

We have introduced a new energy based process representation, \ebp s,
that unifies the stochastic process and latent variable modeling
perspectives for set distributions.
The proposed framework enhances the flexibility of current
process and latent variable approaches, with provable exchangeability and consistency,
in the conditional and unconditional settings respectively.
We have also introduced a new neural collapsed inference procedure
for practical training of \ebp s,
and demonstrated strong empirical results across a range of
problems that involve conditional and unconditional set distribution modeling.
Extending the approach to distributions over
sets of discrete elements remains an interesting direction for
future research.

\subsubsection*{Acknowledgments}
We thank Weiyang Liu, Hongge Chen, Adams Wei Yu, and other members of the Google Brain team for helpful discussions.


\bibliography{reference}
\bibliographystyle{icml2020}


\clearpage
\newpage

\appendix
\onecolumn

\begin{appendix}

\thispagestyle{plain}
\begin{center}
{\huge Appendix}
\end{center}
\numberwithin{equation}{section}

\section{Adversarial Dynamics Embedding}\label{appendix:ade}

The details of ADE derivations were originally provided in~\citep{dai2019exponential}. We include relevant details here for completeness, since we make explicit use of these techniques in our training methods.

Given the EBM with $p_f\rbr{x}\propto \exp\rbr{f\rbr{x}}$, ADE considers the augmented model 
\begin{equation}\label{eq:augmented_model}
	p_f\rbr{x, v}\propto \exp\rbr{f\rbr{x} - \frac{\lambda}{2}v^\top v},
\end{equation}
where $v$ is the auxiliary momentum variable. It has been proved in~\citet{dai2019exponential} that the MLE of~\eqref{eq:augmented_model} is the same as the original model, \ie, 
\begin{equation}
\argmax_{f} \widehat\EE_{\Dcal}\sbr{\log \int p\rbr{x, v} dv} = \argmax_f \widehat\EE_{\Dcal}\sbr{\log p_f\rbr{x}}. 
\end{equation}
We then apply the primal-dual view of the MLE to the augmented model, leading to 
\begin{equation}\label{eq:primal_dual_augmented_mle}
\max_{f}\min_{q\rbr{x, v}}\,\, \widehat\EE\sbr{f\rbr{x}} - \EE_{q\rbr{x, v}}\sbr{f\rbr{x} - \frac{\lambda}{2}v^\top v - \log q\rbr{x, v}},
\end{equation}
where $q\rbr{x, v}$ is the dual sampler in the exponential family of distributions.  

To ensure that the dual sampler $q\rbr{x, v}$ is flexible and tractable, ADE utilizes the dynamics embedding parametrization. Specifically, we consider the Hamiltonian dynamics embedding as an example. In this setting, the sample first comes from an initial distribution $\rbr{x^0, v^0}\sim q_{\omega}^0\rbr{x, v}$, and then moves according to
\begin{equation}\label{eq:leapfrog}
\rbr{x', v'} = \Lb_{f, \eta}\rbr{x,v} \defeq
\rbr{\begin{array}{c}
v^{\frac{1}{2}} = v+ \frac{\eta}{2}\nabla_x f\rbr{x} \\
\quad x' = x + \eta v^{\frac{1}{2}}\\
\quad v' = v^{\frac{1}{2}} + \frac{\eta}{2}\nabla_x f\rbr{x'}
\end{array}},
\end{equation}
where $\eta$ is defined as the leapfrog stepsize. After $T$ iterations, we obtain
\begin{equation}\label{eq:hmc_nn}
\rbr{x^T, v^T} =
\Lb_{f, \eta}\circ\Lb_{f, \eta}\circ\ldots\circ\Lb_{f, \eta}\rbr{x^0, v^0}
,
\end{equation} 
where $\Lb_{f, \eta}$ can be one layer of the neural network. Together with the initial distribution $q^0_{\omega}$, we obtain the parametrization of $q\rbr{x, v}$  with learnable parameters $\rbr{\eta, \omega}$. As justified in~\citet{dai2019exponential}, this parametrization is flexible and has a tractable density,
\begin{equation}
q^T\rbr{x^T, v^T} = q^0_{\omega}\rbr{x^0, v^0}. 
\end{equation}
By plugging this parametrization into~\eqref{eq:primal_dual_augmented_mle}, we obtain the final objective,

\begin{equation}\label{eq:vanilla_hmc_param}
\max_{f\in\Fcal}\min_{\omega, \eta}\ell\rbr{f, \omega}\defeq 
\widehat\EE_\Dcal\sbr{f}-\EE_{\rbr{x^0, v^0}\sim q_\omega^0\rbr{x, v}}\sbr{f\rbr{x^T} -\frac{\lambda}{2}\nbr{v^T}_2^2} - H\rbr{q_\omega^0}.
\end{equation}

\citet{dai2019exponential} also introduces the Langevin and generalized Hamiltonian dynamics embedding for dual density parametrization in ADE. For more details, please refer to~\citet{dai2019exponential}.

\subsection{Dynamics Embedding Distribution Parametrization}\label{appendix:ade_param}

We now present the concrete implementation of the dual sampler $q\rbr{x_{1:n}, v|\theta}$ used in our paper. Following the ADE technique introduced above, parametrization of $q\rbr{x_{1:n}, v|\theta}$ is separated into parameterizing the initial distribution and the dynamics embedding. In our real-data experiment, we use the block RNN to parametrize $q_\omega^0\rbr{x_{1:n}, v|\theta}$, as introduced in~\appref{appendix:exp_arch}. Block RNN is simply a design choice; other alterantives such as normalizing flows can also parametrize $q_\omega^0\rbr{x_{1:n}, v|\theta}$, as shown in our synthetic-data experiment.

The output of the RNN is then treated as the starting sample to which $T$ Hamiltonian/Langevin dynamics updates are applied, \ie, 
\begin{equation}\label{eq:dynamics_nn}
\rbr{x_{1:n}^T, v^T} =
\Lb\circ\Lb\circ\ldots\circ\Lb\rbr{x_{1:n}^0, v^0},
\end{equation} 
where $\Lb$ can be either a Hamiltonian layer or a Langevin layer as specalized below,
\begin{eqnarray}\label{eq:hamiltonian_layer}
\text{\bf Hamiltonian layer:~~}
\rbr{x_{1:n}', v'} = \Lb_{f, \eta}\rbr{x_{1:n},v} \defeq
\rbr{
\begin{array}{c}
v_i^{\frac{1}{2}} = v_i+ \frac{\eta}{2}\nabla_x f\rbr{x_i; \theta} \\
\quad x_i' = x_i + \eta v_i^{\frac{1}{2}}\\
\quad v_i' = v_i^{\frac{1}{2}} + \frac{\eta}{2}\nabla_x f\rbr{x_i';\theta}
\end{array}\Bigg|\,i=1,\ldots, n
},
\end{eqnarray}
with $v' = \cbr{v'_i}_{i=1}^n$.

\begin{eqnarray}\label{eq:langevin_layer}
\text{\bf Langevin layer:~~}
\rbr{x_{1:n}', v'} = \Lb^{\xi}_{f, \eta}\rbr{x_{1:n}} \defeq
\rbr{
\begin{array}{c}
v_i' = \xi_i+ \frac{\eta}{2}\nabla_x f\rbr{x_i; \theta} \\
\quad x_i' = x_i + v_i' \\
\end{array}\Bigg|\,i=1,\ldots, n
},
\end{eqnarray}
with $\xi = \cbr{\xi_i}_{i=1}^n$, $\xi_i\sim q_\omega\rbr{\xi}$ and $v' = \cbr{v'_i}_{i=1}^n$. 

Finally, following Theorem 4 in~\citet{dai2019exponential}, we obtain the parametrized dual sampler with tractable density as 
\begin{eqnarray}
\text{\bf Hamiltonian embedding:~~}&& q^T\rbr{x^T_{1:n}, v^T|\theta} = q^0\rbr{x^0_{1:n}, v^0|\theta},\\
\text{\bf Langevin embedding:~~}&& q^T\rbr{x^T_{1:n}, \cbr{v^t}_{t=1}^T|\theta} = q^0\rbr{x^0_{1:n}, \xi^0|\theta}\prod_{t=1}^{T-1}q_{\omega_i}\rbr{\xi^t}. 
\end{eqnarray}
By plugging this into the primal-dual view objective~\eqref{eq:ebp_obj}, we are able to learn the paramters in {\ebp s} and dual samplers.

\section{Derivation of Special Cases of {\ebp s}}\label{appendix:special_ebp}

In this section, we provide the details for instantiating (un)conditional {\ebp s} to other specific models.

\subsection{Latent Variable Representation of Gaussian Processes}\label{appendix:lvm_gp}

We consider the latent variable model specified in~\eqref{eq:gp_weight}, \ie, 
\begin{eqnarray}
{\theta} &\sim&  \Ncal\rbr{\zero, I_d},\\
f_w\rbr{x, t; \theta} &=& \frac{1}{2\sigma^2}\nbr{x - \theta^\top\phi\rbr{t}}^2.
\end{eqnarray}
To show that the marginal distribution follows
$$
p\rbr{x_{t_1:t_n}|\cbr{t_i}_{i=1}^n} = \Ncal\rbr{0, K\rbr{t_{1:,n}} + \sigma^2 I_n},
$$
we integrate out of $\theta$, \ie, 
\begin{eqnarray*}
p\rbr{x_{t_1:t_n}|\cbr{t_i}_{i=1}^n} &=& \int \prod_{i=1}^n p\rbr{x|t_i, \theta}p\rbr{\theta} d\theta =  \Ncal\rbr{0, \sigma^2 I_n + \phi\rbr{{t_{1:n}}}^\top\phi\rbr{t_{1:n}}},
\end{eqnarray*}
where the last equation come from the integration of Gaussians. This shows that {\gp s} under the latent variable parametrization are a special case of {\ebp s}.

\subsection{Latent Variable Representation of Student-$t$ Processes}\label{appendix:lvm_tp}

We consider the latent variable model specified in~\eqref{eq:tp_weight}, \ie, 
\begin{eqnarray}
\alpha&\sim& \Ncal\rbr{\zero, I_d},\\
\beta^{-1} &\sim& \Gamma\rbr{\frac{\nu}{2}, \frac{\gamma}{2}},\\
f_w\rbr{x, t;\theta} &=& \frac{\gamma\nbr{x - \sqrt{\frac{\beta\rbr{\nu - 2}}{\gamma}}\alpha^\top \phi\rbr{t}}^2}{2{\sigma^2\rbr{\nu - 2}\beta}},
\end{eqnarray}
To show the marginal distribution follows
\begin{eqnarray*}
p\rbr{x_{t_1:t_n}|\cbr{t_i}_{i=1}^n} = \Tcal\rbr{\nu, 0, K\rbr{t_{1:n}} + \beta I_n},
\end{eqnarray*}
we integrate out of $\theta =\rbr{\alpha, \beta}$. By integrating over $\alpha$, we have
\begin{eqnarray*}
p\rbr{x_{t_1:t_n}|\beta,\cbr{t_i}_{i=1}^n} &=& \int \prod_{i=1}^n p\rbr{x_{t_i}|t_i, \alpha, \beta} p\rbr{\alpha}d\alpha\\
&=& \Ncal\rbr{\zero, \rbr{\frac{\beta}{\gamma} \rbr{\nu - 2}} \rbr{\underbrace{\sigma^2 I_n + \phi\rbr{t_{1:n}}^\top\phi\rbr{t_{1:n}}}_{\Ktil(t_{1:n})} } },
\end{eqnarray*}
and by integrating over $\beta$, we have
\begin{eqnarray*}
p\rbr{x_{t_1:t_n}|\cbr{t_i}_{i=1}^n} &=& \int p\rbr{x_{t_1:t_n}|\beta, \cbr{t_i}_{i=1}^n}p\rbr{\beta} d\beta\\
&\propto&\int \exp\rbr{ - \frac{1 + \frac{ \gamma X^\top \Ktil^{-1} X}{\rbr{\nu-2}}} {2\beta}} \beta^{ - \frac{\nu}{2} - 1} d\beta\\
& \propto &  \rbr{1 + \frac{ \gamma X^\top \Ktil^{-1} X}{\rbr{\nu-2}}}^{\frac{\nu + n}{2}} = \Tcal\rbr{\nu, 0, \Ktil}
\end{eqnarray*}
This shows that {\tp s} under the latent variable parametrization are a special case of {\ebp s}.

\subsection{Topic Models}\label{appendix:lvm_topic}

The topic models, including Bayesian sets~\citep{GhaHel05}, probabilistic latent semantic index~\citep{Hofmann99}, latent Dirichlet allocation~(LDA) family~\citep{BleNgJor03,BleLaf07,BleMcA07}, and replicated softmax~\citep{hinton2009replicated}, are also special cases of the unconditional {\ebp s}. 

Here we consider the original LDA and replicated softmax as examples of directed and undirected topic models respectively. Other models follow a similar consideration.

\paragraph{Latent Dirichlet Allocation} LDA is a representative of the directed topic model, which treats a document as a set of words. The model defines each component in~\eqref{eq:ebp_unconditional} as
\begin{eqnarray*}\label{eq:lda}
p\rbr{t_i} &=& \Dcal ir\text{-}\Mcal ulti\rbr{\alpha},\\
p\rbr{x_{t_i}|t_i} &=& \Mcal ulti\rbr{\Phi^\top \one_{t_i}}, \quad i = 1,\ldots, n,
\end{eqnarray*}
where $t_i\in \cbr{1,\ldots, d}$ denotes the topic of each word and follows the Dirichlet-Multinomial distribution. The $x_i$ is a $k$-dimensional one-hot encoding for each word in the vocabulary. $\Phi$ is a $d\times k$ matrix where each row denotes the distribution of words in one topic.

\paragraph{Replicated Softmax} The replicated softmax~\citep{hinton2009replicated} is proposed as an undirected topic model, which defines the joint distribution in~\eqref{eq:ebp_unconditional} as a restricted Boltzmann machine with a linear potential function, \ie, 
\begin{eqnarray}\label{eq:replicated_softmax}
p\rbr{x_{1:n}, \theta} \propto \exp\rbr{- \sum_{i=1}^n \rbr{\theta^\top W x_i + b^\top x_i } - a^\top\theta},
\end{eqnarray}
where $\theta\in \{0, 1\}^d$ can be seen as the latent topic assignment. Each $x_i$ is a $k$-dimensional indication vector with only one element equals to $1$ and the rest equal to $0$. Together $\cbr{x_i}_{i=1}^n$ denote the one-hot encodings for the observed words in the document.

As a result, LDA and replicated softmax are special realizations of uncondtional {\ebp s}, where the index variables are either explicit defined or integrated out.

\section{Proof Details of~\thmref{thm:prior_consistent}}\label{appendix:proof_prior}

{\bf~\thmref{thm:prior_consistent}}
\textit{If $n\ge m\ge 1$, and prior is exchangeable and consistent, then the marginal distribution $p\rbr{x_{1:n}}$ will be exchangeable and consistent. 
} 
\begin{proof}
We simply verify the consistency of $p\rbr{x_{1:n}}$ under the consistency of $
p\rbr{\cbr{t_i}_{i=1}^m}$. 
\begin{eqnarray*}
&&\int p\rbr{x_{1:n}}dx_{m+1:n}\\
& = &\int \int p\rbr{x_{t_1:t_n}|\cbr{t_i}_{i=1}^n} p\rbr{\cbr{t_i}_{i=1}^n} d\cbr{t_i}_{i=1}^n dx_{m+1:n}\\
&=&\int \rbr{\int p\rbr{x_{t_1:t_n}|\cbr{t_i}_{i=1}^n}  dx_{m+1:n} } p\rbr{\cbr{t_i}_{i=1}^n}  d\cbr{t_i}_{i=1}^n\\
&=&\int p\rbr{x_{t_1:t_{m}}|\cbr{t_i}_{i=1}^m} \rbr{\int p\rbr{\cbr{t_i}_{i=1}^n}  d t_{m+1:n}} d t_{1:m}\\
& = & \int p\rbr{x_{t_1:t_{m}}|\cbr{t_i}_{i=1}^m} p\rbr{\cbr{t_i}_{i=1}^m} d t_{1:m} = p\rbr{x_{1:m}}.
\end{eqnarray*}
The exchangeability of $p\rbr{x_{1:n}}$ directly comes from the exchangeablity of $p\rbr{x_{t_{1}:t_n}}$ and $p\rbr{\cbr{t_i}_{i=1}^n}$,
\begin{eqnarray*}
p\rbr{x_{1:n}} &=& \int p\rbr{x_{t_1:t_n}|\cbr{t_i}_{i=1}^n} p\rbr{\cbr{t_i}_{i=1}^n} d\cbr{t_i}_{i=1}^n \\
&=&\int p\rbr{\pi\rbr{x_{t_1:t_n}}|\pi\rbr{\cbr{t_i}_{i=1}^n}} p\rbr{\pi\rbr{\cbr{t_i}_{i=1}^n}} d\cbr{t_i}_{i=1}^n \\
&=& p\rbr{\pi\rbr{x_{t_1:t_n}}}
\end{eqnarray*}
\end{proof}

\section{More Details of Inference}\label{appendix:inference_details}

\subsection{Further Neural Collapsed Inference for Unconditional {\ebp s}}\label{appendix:ebp_further_inference}

In the main text, we introduce the neural collapsed inference for {\ebp s} to eliminate the posterior inference of $\cbr{t_i}_{i=1}^n$. We can further exploit the neural collapsed inference idea to eliminate $\theta$. 

Recall the model with $\cbr{t_i}_{i=1}^n$ collapsed,
\begin{eqnarray*}
p_{w'}\rbr{x_{1:n}|\theta} &=& \prod_{i=1}^n \int p_w\rbr{x_{t_i}|\theta, t_i}p\rbr{t_i} dt_i\\
&&\propto \prod_{i=1}^n \int {\exp\rbr{f_w\rbr{x_{t_i}, t_i; \theta} - Z\rbr{f_w, t_i;\theta} + h_v\rbr{t_i}}} dt_i\\
&&\approx \prod_{i=1}^n \frac{1}{Z\rbr{f_{w'}; \theta}}\exp\rbr{f_{w'}\rbr{x_i;\theta}}.
\end{eqnarray*}
We further consider $p\rbr{\theta}\propto \exp\rbr{g_u\rbr{\theta}}$, then we have
\begin{eqnarray}\label{eq:further_nci}
&&\prod_{i=1}^n \int p_{w'}\rbr{x_{1:n}|\theta}p\rbr{\theta}d\theta\nonumber \\
&\propto& \int \exp\rbr{\sum_{i=1}^n f_{w'}\rbr{x_i; \theta} + g_u\rbr{\theta}} d\theta \nonumber \\
&\approx& \frac{1}{Z\rbr{{w''}}}\exp\rbr{\sum_{i=1}^n f_{\wtil'}\rbr{x_i; \sum_{i=1}^n \phi_{u'}\rbr{x_i}}}\defeq p_{w''}\rbr{x_{1:n}},
\end{eqnarray}
where we denote $w'' = \cbr{\wtil', u'}$. The last approximation comes from the fact that
\begin{itemize}
	\item[1)] the integration over $\theta$ leads to a distribution over $x_{1:n}\in \otimes^n \Omega$;
	\item[2)] since $p\rbr{x_{1:n}}$ is an integration of latent variable model, it should be exchangeable. 
\end{itemize}
Therefore, we consider deepsets~\citep{zaheer2017deep} in the reparametrized EBM as a special choice of~\eqref{eq:further_nci}, which satifies these two conditions in order to approximate the integrated model. 

The attention mechanism is another choice for the reparametrized EBM, \ie, 
$$
p_{w''}\rbr{x_{1:n}} =  \frac{1}{Z\rbr{{w''}}}\exp\rbr{\sum_{i=1}^n\mathtt{attn}_{\wtil'}\rbr{x_i; {x_{1:n}}}},
$$
where $\mathtt{attn}_{w''}\rbr{x; x_{1:n}} = \sum_{i=1}^n\frac{\exp\rbr{\phi\rbr{x}^\top\phi\rbr{x_{i}}}h\rbr{x_i}}{\sum_{j=1}^n \exp\rbr{\phi\rbr{x}^\top\phi\rbr{x_j}}}$ with $w''$ denoting the parameters in $\phi\rbr{\cdot}$ and $h\rbr{\cdot}$.

We can apply ADE to such neural collapsed reparametrization if the task only concerns set generation (i.e., does not use $\theta$). We take the deepsets parametrization~\eqref{eq:further_nci} as an example (the attention parametrization follows similarly). Specifically, we learn the parameters in $w''$ via
\begin{multline}\label{eq:further_nci_ade}
\max_{w''}\min_{q\rbr{\cbr{x_i}_{i=1}^n, v}}\,\, \widehat\EE_{\Dcal}\sbr{\sum_{i=1}^n f_{\wtil'}\rbr{x_i; \sum_{i=1}^n \phi_{u'}\rbr{x_i}}}\\
 - \EE_{q\rbr{\cbr{x_i}_{i=1}^n, v}}\sbr{\sum_{i=1}^n f_{\wtil'}\rbr{x_i; \sum_{i=1}^n \phi_{u'}\rbr{x_i}} - \frac{\lambda}{2}v^\top v} - H\rbr{q\rbr{\cbr{x_i}_{i=1}^n, v}}. 
\end{multline}
Following the ADE technique, we can parametrize the initialization distribution $q_{\omega}^0\rbr{\cbr{x_i^0}_{i=1}^n, v^0}$ using an RNN and refine $q\rbr{\cbr{x_i}_{i=1}^n, v}$ with learnable Hamiltonian/Langevin dynamics as introduced in~\secref{appendix:ade}.

\paragraph{Connection to Gibbs Point Processes~(\gpp s) and Determinantal Point Processes~(\dpp s):} In fact, if we adopt the most general model collapsed unconditional~\ebp~for arbitrary $n$, we obtain
\begin{equation}\label{eq:general_nci}
p_{w''}\rbr{x_{1:n}} = \frac{1}{Z_{w''}} \exp\rbr{f_{w''}\rbr{x_{1:n}}},
\end{equation}
which includes the Gibbs point processes~\citep{Dereudre19} with $f_{w''}$ satisfying several mild conditions for the regularity of~\gpp. Particularly, the determinatal point processes~\citep{lavancier2015determinantal,kulesza2012determinantal} can be instantiated from~\gpp~by setting the potential function to be $\log\det$ of some kernel function, which can also be parametrized by neural network, \eg,~\citet{xie2017deep}.

Therefore, the proposed algorithm can straightforwardly applied for (deep)~\gpp s and \dpp s learning. It should be emphasized that by exploiting the proposed primal-dual MLE framework, we automatically obtain a deep neural network parametrized dual sampler with the learned model simultaneously, which can be used in inference and bypass the notorious sampling difficulty in \gpp~and \dpp.

\subsection{Inference for Conditional {\ebp}}\label{appendix:ebp_cond_inference}

In the conditional {\ebp} setting where the index $\cbr{t_i}_{i=1}^n$ are given for any cardinality $n$, we are modeling:
\begin{equation}
  p\rbr{x_{t_1:t_n}|\cbr{t_i}_{i=1}^n} = \int \prod_{i=1}^{n} p(x | \theta, t_i) p(\theta)d\theta
\end{equation}
For simplicity, we use capital letters to denote the set. We denote $T_{train}, X_{train}$ as the observed sets of points, and $T_{test}, X_{test}$ as the un-observed inputs and targets. For the rest of this section, we denote $T = T_{train} \cup T_{test}$ and $X = X_{train} \cup X_{test}$. Then the predictive model is to infer
\begin{equation}
	p(X_{test} | T, X_{train}) = \frac{p(X_{train}, X_{test} | T)}{\sum_{X'}p(X_{train}, X' | T)} = \frac{p(X_{train}, X_{test} | T)}{p(X_{train} | T)} = \frac{p(X_{train}, X_{test} | T)}{p(X_{train} | T_{train})}
	\label{eq:pred_bayes}
\end{equation}
where the last equal sign is due to the consistency condition of stochastic processes. Below we briefly review several existing conditional neural processes. 
\subsubsection{Review of conditional neural processes}

The learning of processes is generally done via maximizing the marginal likelihood $p(X|T)$. The following neural processes, however, perform learning by maximizing the predictive distribution $p(X|T, X_{train})$ directly.

\noindent\textbf{Conditional Neural Processes ~\citep{garnelo2018conditional}} directly parametrize the conditional distribution $p(X|T, X_{train})$ as: 
\begin{equation}
	p(X|T, X_{train}) = \prod_{i=1}^{|X|} \Ncal(x_{t_i} | \mu_{t_i}, \sigma_{t_i})
\end{equation}
where $\mu_{t_i}$ and $\sigma_{t_i}$ are outputs of some neural network $g(t_i, T_{train}, X_{train})$ with permutation invariance. 

\noindent\textbf{Neural Processes ~\citep{garnelo2018neural}} model the distribution of $p(X|T)$ as
\begin{equation}
	p(X|T) = \int \prod_{i=1}^{|X|} \Ncal(x_{t_i} | g(t_i, \theta), \sigma) p(\theta)d\theta,
\end{equation}
where $g(t_i, \theta)$ can be learned using ELBO:
\begin{equation}
	\log p(X|T) \geq  \EE_{q(\theta|T, X)} \Big[ \sum_{i=1}^{|X|} \log \Ncal(x_{t_i} | g(t_i, \theta), \sigma) + \log p(\theta) - \log q(\theta|T, X) \Big].
\end{equation}
During learning, however,~\citet{garnelo2018neural} again performs MLE on the predictive model:
\begin{equation}
	\log p(X_{pred} | T, X_{context}) \geq  \EE_{q(\theta|T, X)} \Big[\sum_{x_{t_j} \in X_{pred}}  \log \Ncal(x_{t_j} | g(t_j, \theta), \sigma) 
	+ \log p(\theta|T_{context}, X_{context}) - \log q(\theta|T, X) \Big],
\end{equation}
where $X_{context}$ denotes a subset of $X_{train}$ and $X_{pred} = X_{train}\setminus X_{context}$, that are randomly splitted from $X_{train}$. The $p(\theta|T_{context}, X_{context})$ is the true posterior after observing $T_{context}, X_{context}$. This posterior then serves as the prior of the predictive model according to Bayes' rule. However, since $p(\theta|T_{context}, X_{context})$ is not tractable in general, ~\citet{garnelo2018neural} uses $q(\theta|T_{context}, X_{context})$ to approximate this term instead. 

\noindent\textbf{Attentive Neural Processes~\citep{kim2019attentive}} is an extension to Neural Processes~\citep{garnelo2018neural} by using self-attention to parametrize the variational posterior. Besides $\theta$ from the variational posterior, their model also adds the deterministic context embedding $r$ computed using attention, together with location $x_{t_i}$, into the mean function $g(x_{t_i}, r, \theta)$. Other than these differences, the training procedure is the same as~\citet{garnelo2018neural}, which still uses Gaussian for observation modeling.

\subsubsection{Learning conditional {\ebp}}

The proposed {\ebp s} exploits flexible energy-based model, intead of Gaussian distributions in {\np s} and theirs variants:
\begin{equation}
	p(X | T, \theta) \propto \exp\rbr{f(X, T; \theta)} p(\theta) \label{eq:our_1d_model} \propto \exp\rbr{\sum_{i=1}^{|X|} f(x_{t_i}, t_i;\theta) }.
\end{equation}
The ELBO then becomes:
\begin{equation}
	\log p(X|T) \geq  \EE_{q(\theta|T, X)} \Big[ \sum_{i=1}^{|X|} f(x_{t_i}, t_i; \theta)) - A(\theta, t_i) 
	+ \log p(\theta) - \log q(\theta|T, X) \Big],
\end{equation}
where $A(\theta, t) = \log \int \exp(f(x, t; \theta))dx$ is the log partition function. 

Using the ADE technique, we have $A(\theta, t) =\max_{q'\rbr{x|\theta, t}} \EE_{q'(x|\theta, t)} \sbr{f(x, t; \theta)} + H(q')$. By plugging this into the ELBO, we arrive at the learning objective which tries to maximize the marginal likelihood w.r.t.
\begin{equation}
	\max_{q, h}\min_{q'} \EE_{q(\theta|T, X)} \Big[  \log p(\theta) - \log q(\theta|T, X) + \sum_{i=1}^{|X|} f(x_{t_i}, t_i; \theta))
	- \EE_{x \sim q'} f(x, t_i; \theta) - H(q') 
	 \Big].
\end{equation}

Different from the family of neural processes, we are able to directly optimize for $p(X|T)$ using the above objective. We can easily learn the predictive distribution $p(X|T, X_{train})$ similar to other neural processes by replacing the prior with the variational posterior from the observed set $T_{train}, X_{train}$.

\subsubsection{Prediction using {\ebp s}}

Without loss of generality, we illustrate the prediction of a single point $(t, x)$ given the training set, namely $p(x|t, X_{train}, T_{train})$. 

As is pointed out in Eq~\eqref{eq:pred_bayes}, $p(x|t, X_{train}, T_{train}) \propto p(x, X_{train} |t, T_{train})$. We use the variational posterior (denoted as $q(\theta)$ below for simplicity) to approximate the lower bound:
\begin{eqnarray}
	\log p(x, X_{train} |t, T_{train}) &=& \log \rbr{ \int p(\theta) \exp\rbr{f(x, t; \theta) - A(\theta, t)} \prod_{i=1}^{|X_{train}|} \exp\rbr{f(x_{t_i}, t_i; \theta)) - A(\theta, t_i)}d\theta } \nonumber \\
	&\geq & \max_{q(\theta)} \EE_{q(\theta)}\sbr{f(x, t; \theta)} + \EE_{q(\theta)} \sbr{G(\theta, t, X_{train}, T_{train})},
\end{eqnarray}
where
\begin{equation}
	G(\theta, t, X_{train}, T_{train}) = \log p(\theta) - \log q(\theta) - A(\theta, t) + \sum_{i=1}^{|X_{train}|} f(x_{t_i}, t_i; \theta)) - A(\theta, t_i) 
\end{equation}
In practice, we use $q(\theta|X_{train} |T_{train})$ to replace the above $p(\theta)$, which results in the predictive distribution approximated by:
\begin{eqnarray}
\label{eq:pred_approx}
	p(x|t, T_{train}, X_{train}) &\geq &  \exp\rbr{\EE_{q(\theta)}\sbr{f(x, t; \theta)}} \cdot \exp\rbr{\EE_{q(\theta)}G(\theta, t, X_{train}, T_{train})} \nonumber \\
	& \propto & \exp\rbr{\EE_{q(\theta)}\sbr{f(x, t; \theta)}} \nonumber \\ 
                               & \simeq & \exp\rbr{\EE_{q(\theta|X_{train} |T_{train})}\sbr{f(x, t; \theta)}}                                          
\end{eqnarray}

\section{Additional Related Work}\label{appendix:related_work}

The proposed~{\ebp s} bridge the gap between stochastic processes and models of exchangeable data. We summarize these two related topics below:

\paragraph{Stochastic Processes.} Exploiting stochastic processes for conditional distribution modeling has a long line of research, starting from Gaussian processes for regression~\citep{WilRas96} to being generalized to classification~\citep{OppWin00b} and ordinal regression~\citep{ChuGha05}. One of the major bottlenecks of {\gp s} is the memory and computational costs --- $\Ocal\rbr{N^2}$ in memory and $\Ocal\rbr{N^3}$ in computation respectively, where $N$ denotes the number of training samples. Although low-rank and sparse approximations have been proposed~\citep{WilSee00b,QuiRas05,SneGha07,titsias2009variational,HenFusLaw13} to reduce these costs, flexibilty of {\gp s} and its variants remains a critical issue to be addressed. 

Student-$t$ processes are derived by adding the inverse-Wishart process prior to the kernel function~\citep{shah2014student}, allowing the kernel function to adapt to data. As we discussed in~\secref{subsec:ebp_construction}, student-$t$ processes essentially rescales each dimention in the likelihood model, and thus still has restricted modeling flexibilty. Another line of research to further improve the flexibility of {\gp s} is to introduce structures or deep compositions into kernels, such as~\citet{duvenaud2013structure,DamLaw12,bui2016deep,wilson2016stochastic,al2017learning}. Even though neural networks can be incorporated in constructing these kernels, the likelihood estimations of these models are still restricted to known distributions, limiting their modeling flexibility.

Neural processes~\citep{garnelo2018neural} and its varaints~\citep{garnelo2018conditional,kim2019attentive,louizos2019functional} introduce neural networks to stochatic processes beyond {\gp s}. However, their likelihoods are still restricted to known distributions, \eg, Gaussian. Besides the modeling restriction, these models are learned by maximizing $\log$-\emph{predictive} distribution, rather than the $\log$-\emph{marginal} distibution, which may lead to suboptimial solutions.

The most competitive model w.r.t. {\ebp s} is the implicit processes~\citep{ma2018variational}, which uses the implicit models as the likelihood in~\eqref{eq:sp_lvm}. However, due to the intractability of the implicit likelihood,  {\gp s} are introduced for variational inference, which negatively impacts the initially designed flexility in implicit processes, as demonstrated in~\secref{sec:appx_syn}.

\paragraph{Exchangeable Probabilistic Models.} The generative models for exchangeable data is a separate research topic which mostly relies on the De Finetti's Theorem. Bayesian sets~\citep{GhaHel05} considers the latent variable model with Bernoulli distribution likelihood and beta prior for a set of binary data. The conjugacy of Bernoulli and beta distributions leads to tractability in Bayesian inference, but limits the flexibility. Topic models, \eg, pLSI~\citep{Hofmann99}, LDA family~\citep{BleNgJor03,BleLaf07,BleMcA07}, and replicated softmax~\citep{hinton2009replicated} generalize Bayesian sets by introducing more complicated local latent variables, but the likelihood is still restricted to simple distributions. 

To make distribution modeling more flexible, neural networks have been introduced to likelihood estimation of latent variable models~\citep{edwards2016towards}. However, this work only exploits the analytic form of parametrization within known distributions. \citet{KorDegHusGaletal18,yang2019pointflow} use normalizing flows to improve modeling flexibility of exchangeable data. Flow-based models, compared to {\ebp s}, still restrict the underlying distribution (by requiring the transformations to be invertable), and therefore cannot fully utilize the expressiveness of neural networks. 

The Gibbs point processes~\citep{Dereudre19}, including Poission point processes, Hawkes point processes~\citep{Hawkes71} and determinantal point processes~\citep{lavancier2015determinantal,kulesza2012determinantal} as special cases, is also an alternative flexible model for exchangeable data. As we discussed in~\appref{appendix:ebp_further_inference}, by the neural collapsed inference technique, we reduce the collapsed~\ebp~to~\gpp, which implies the flexibility of the proposed~\ebp. Moreover, this connection also highlights that the proposed primal-dual ADE can be used for~\gpp~and~\dpp~estimation.

\section{Experiment Details}\label{appendix:exp_details}
\subsection{Architecture and Training Details}\label{appendix:exp_arch}
As preluded in~\secref{appendix:inference_details}, we use deepsets~\citep{zaheer2017deep} to parametrize $q(\theta|x_{1:n})$ as a diagonal Gaussian. Specifically, each input $x_i$ undergoes 1D-convolutions with kernel size 1 and filter sizes $\cbr{128, 256}$  for synthetic-data experiments and filter sizes $\cbr{128, 256, 256, 512}$ for image completion and point-cloud generation. The 1D-convolution layers are interleaved with ReLU, followed by max-pooling across inputs in the embedding space. We then use the reparametrization trick~\citep{KinWel13} to sample $\theta$ from the resulting Gaussian. To compute the energy of a single input $f(x_i;\theta)$, each $x_i$ is concatenated with $\theta$ and followed by fully connected layers of $\cbr{128, 64, 1}$ neurons each interleaved with ReLU. The energy of the set, $f(x_{1:n};\theta)$, is the average energy of all inputs in the set.

To obtain the ADE initialization distribution $q^0\rbr{x_{1:n}^0|\theta}$ on synthetic data, we first use a 2-layer hyper-network~\citep{HaDaiLe17} with 256 hidden neurons each to output the parameters for a 10-layer normalizing planar flow model~\citep{rezende2015variational}. The planar flow then takes both $\theta$ (learned from $T_{train}$ and $X_{train}$) and the target indices $T_{test}$ as inputs to produce $q^0\rbr{x_{t_1:t_n}^0|\theta}$. For the synthetic experiments, we directly use $q^0\rbr{x_{1:n}^0|\theta}$ as the sampler output without performing additional Hamiltonian/Langevin dynamics, which is sufficient to capture the synthetic data distributions.

To obtain $q\rbr{x_{1:n}, v|\theta}$ for image completion and point-cloud generation, we first parametrize the initialization distribution $q^0\rbr{x_{1:n}^0|\theta}$ using an RNN where each recurrent LSTM block consists of an MLP with $\cbr{64, 128, 512}$ hidden neurons interleaved with ReLU. Each LSTM block outputs the mean and variance of a diagonal Gaussian of dimension $k$ times $d$, where $k$ (block size) is the number of elements generated at once, and $d$ is the dimension of each element. For image completion, $k$ equals the number of pixels in a row of an image (\eg, $28$ for MNIST and $32$ for CelebA), and $d$ equals the number of channels of a pixel ($1$ for MNIST and $3$ for CelebA). For point-cloud generation, we have $k=512$ and $d=3$. We considered a range of $k$ values from $32$ to $2048$, and selected $k$ based on the best completion/generation performance. The number of recurrent RNN blocks equals the total number of elements in the final generated set ($784$ for MNIST, $1024$ for CelebA, and $2048$ for point clouds) divided by $k$. Using the RNN output as the initialization distribution $q^0\rbr{x_{1:n}^0|\theta}$, we then perform $T = 20$ steps of Langevin dynamics with step size $\eta=0.1$ and $\xi_i\sim \Ncal\rbr{0, 0.05}$ while clipping $\nabla_x f$ to 0.1.

We use spectral normalization in training the energy function and batch normalization in training the sampler. We use Adam with learning rate $10^{-4}$ to optimize all of our models. We set $\beta_1=0.5$ for image completion and $\beta_1=0.0$ for the synthetic experiments and for point-cloud generation. The coefficient of $H(q\rbr{x_{1:n}, v|\theta})$ in~\eqref{alg:ebp_mle} is set to $10^{-5}$. All tasks are trained until convergence using batch size $64$ on a single NVIDIA V100 GPU with $32$ GB memory.

\subsection{Point-Cloud Preprocessing}
Following~\citet{achlioptas2017learning}, we use shapes from ShapeNet~\citep{WuSonKhoYuetal15} that are axis aligned and centered into the unit sphere. For the point-cloud classification task, we apply random rotations along the gravity axis following~\citet{achlioptas2017learning,yang2019pointflow}, and also normalize shapes from ModelNet40~\citep{WuSonKhoYuetal15} the same way. For generation, the model is trained on the official training split and evaluated on the test split of ShapeNet (where the train-validation-test splits are 70\%-20\%-10\%) similar to~\citet{yang2019pointflow}.

\subsection{Point-Cloud Model Selection}
We follow the model selection protocol of~\citet{achlioptas2017learning}, namely select the model with the smallest JSD and reports other measurements of this model. Measurements are created every 100 epochs. We noticed that the measurement of JSD and COV are relatively robust across different evaluation runs, whereas the MMD measurement (due to its small magnitude, \eg, $10^{-4}$) is less robust.

\subsection{Point-Cloud Generation Metrics}
Following~\citet{achlioptas2017learning,yang2019pointflow}, we use Chamfer distance (CD) and earth mover's distance (EMD) defined below to measure distance between point clouds.

\begin{equation}
  d_{CD}(X, X') = \sum_{x_i\in X} \min_{x_i'\in X'} \|x_i-x_i'\|_2 + \sum_{x_i'\in X'}\min_{x_i \in X} \|x_i-x_i'\|_2,
\end{equation}

\begin{equation}
  d_{EMD}(X, X') = \min_{\phi: X\to X'} \sum_{x_i\in X} \|x_i-\phi(x_i)\|_2.
\end{equation}

Similarly, we use Jensen-Shannon Divergence (JSD), Minimum matching distance (MMD), and Coverage (COV) defined below to evaluate generation quallity.
\begin{itemize}
\item{JSD is computed between the marginal distribution of the entire reference set ($p_r$) and the marginal distribution of the entire generated set ($p_g$) of point clouds, specifically,
\begin{equation}
	\text{JSD}(p_g,p_r) = \frac{1}{2}D_{KL}(p_r||p_m) + \frac{1}{2}D_{KL}(p_g||p_m)\,,  
\end{equation}
where $p_m=\frac{1}{2}(p_r + p_g)$ and $D_{KL}$ stands for the Kullback-Leibler divergence. JSD only measures similarity at the marginal distribution level, and does not provide insights on the generation quality of each individual point cloud.
}
\item{MMD measures the average distance between a point-cloud in the reference set $S_r$ and its closest neighbor in the generated set $S_g$, namely
\begin{equation}
	\text{MMD}(S_g, S_r) = \frac{1}{|S_r|}\sum_{X\in S_r} \min_{X'\in S_g} d(X,X'),
\end{equation}
where $d(X, X')$ is the distance between two point clouds according to either CD or EMD.
}
\item{COV measures the ratio of point clouds in the reference set $S_r$ that are matched to a distinct closest neighbor in the the generated set $S_g$:
\begin{equation}
	\text{COV}(S_g, S_r) = \frac{1}{|S_r|}|\{\arg\min_{X \in S_r} d(X,X') | X' \in S_g \}|.
\end{equation}
where $d(X, X')$ can again be based on CD or EMD. COV reflects the generation diversity.
}
\end{itemize}

\clearpage
\section{More Experimental Results}\label{appendix:more_exp_results}

\subsection{Additional Comparisons on Synthetic Data}\label{sec:appx_syn}

We comapre the proposed {\ebp s} with Gaussian processes~(\gp s), neural processes~(\np s)\footnote{\url{https://github.com/deepmind/neural-processes}}, and variational implicit processes~(\vip s)\footnote{\url{
https://github.com/LaurantChao/VIP}}. 

\begin{figure}[h!]
	\centering
	\includegraphics[width=.8\linewidth]{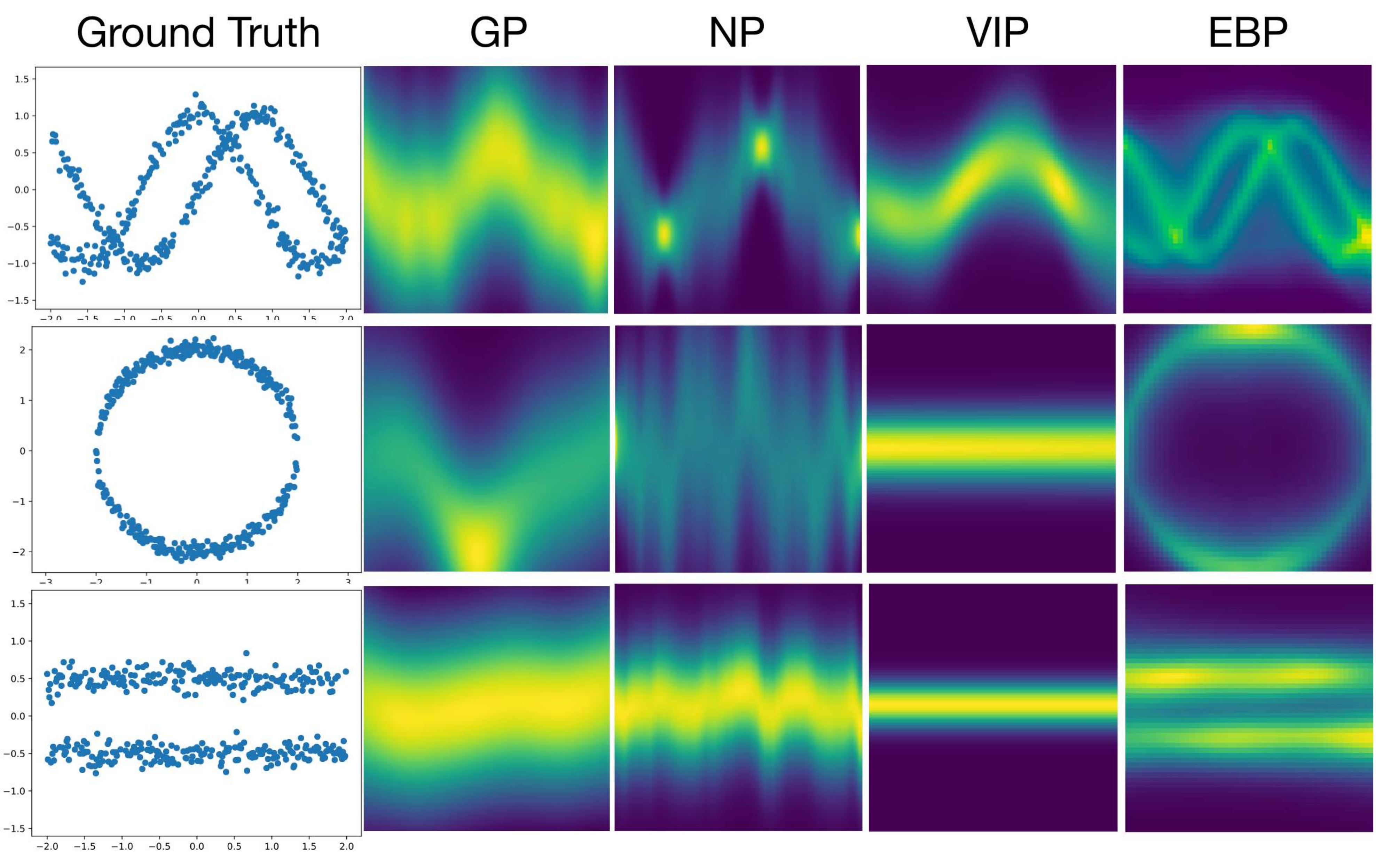}
	\caption{The ground truth data and learned energy functions of {\gp}, {\np}, {\vip}, and {\ebp} (from left to right). {\ebp} successfully captures multi-modality of the toy data as {\gp}, {\np}, and {\vip} exhibiting only a single mode.}
  \label{fig:appx_syn}
\end{figure}

\figref{fig:appx_syn} shows the ground truth data and the normalized probability heatmap for each process. To generate data given an index $t_i$, we randomly select one of the two modes (\eg, one of the sine waves) as the mean and adds $\epsilon \sim \Ncal(0, 0.1)$ noise to produce $x_{t_i}$. 
We plot the heatmap of the learned predictive distribution of each process. For \ebp, we use~\eqref{eq:pred_approx} to approximate $p(x_{t_i}|t_i)$.

\clearpage
\subsection{Additional Image Completion Results on MNIST}
\begin{figure}[h!]
\centering
  \includegraphics[width=.5\linewidth]{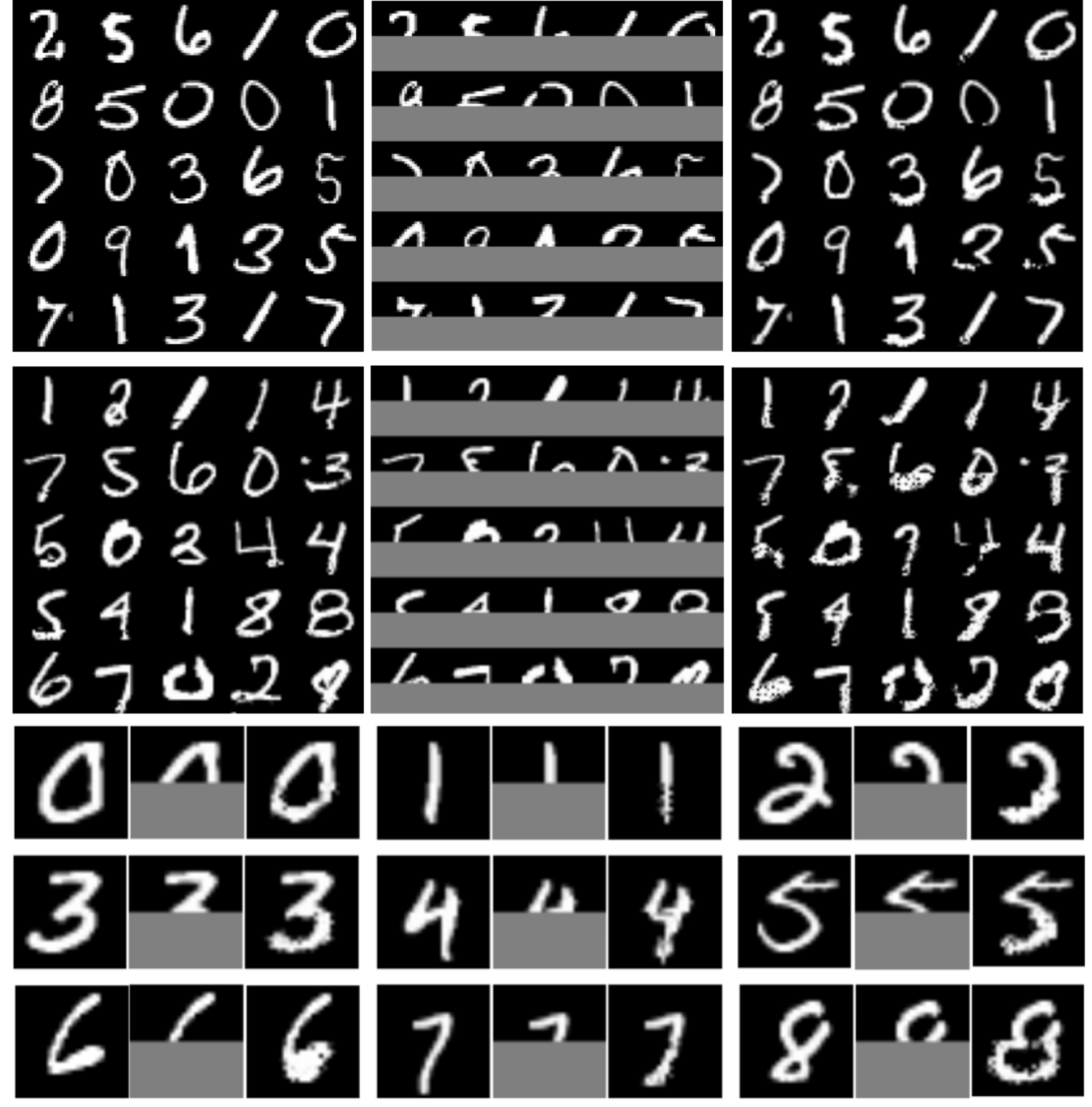}

\caption{Additional image completion results on MNIST where the top half of the image serves as context.}
\label{fig:additional_mnist_1}
\end{figure}

\begin{figure}[h!]
\centering
  \includegraphics[width=.5\linewidth]{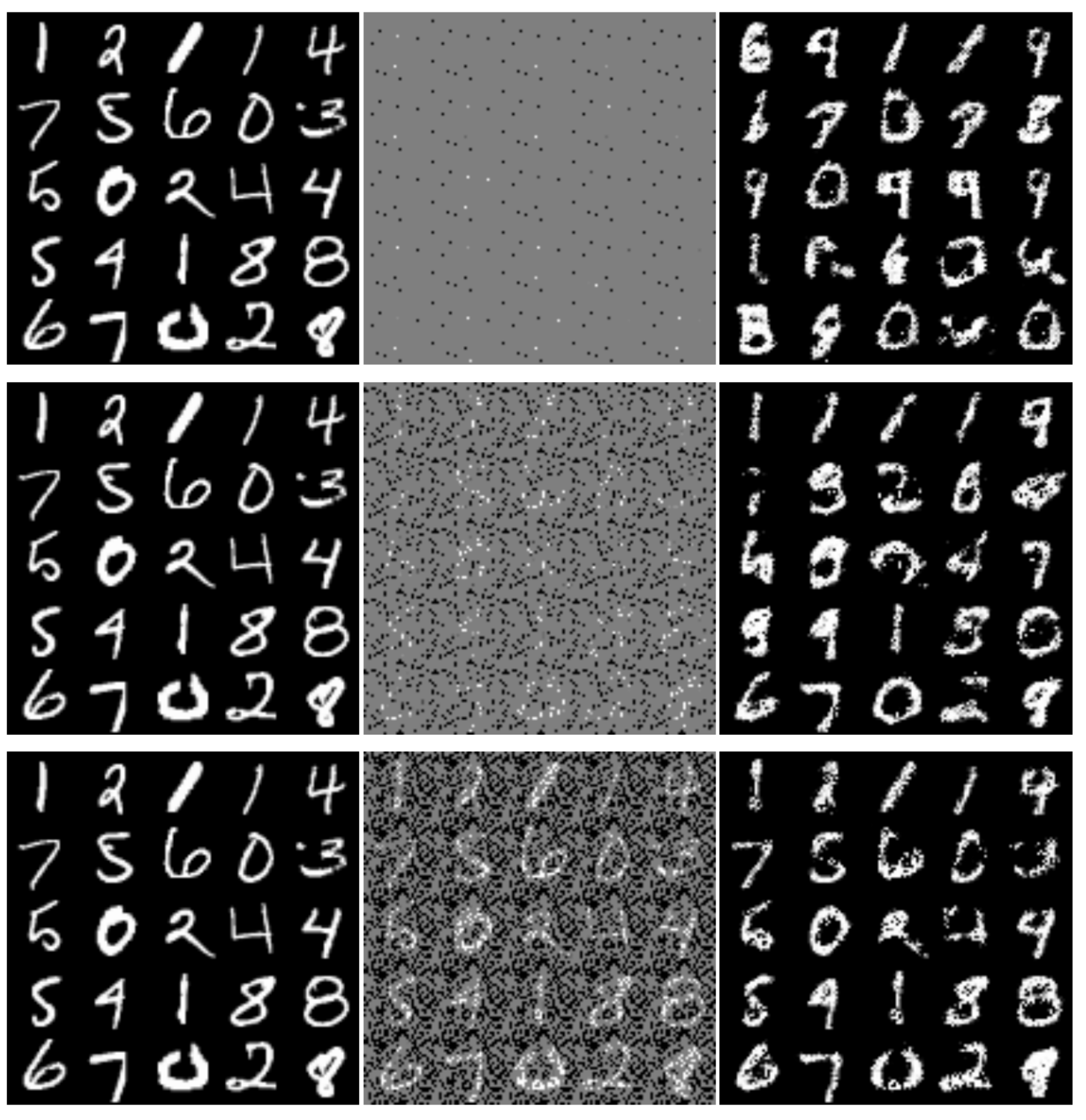}

\caption{Additional image completion results on MNIST where 10, 100, and 1000 (top to bottom) random pixels serve as context.}
\label{fig:additional_mnist_2}
\end{figure}

\figref{fig:additional_mnist_1} shows additional image completion results on MNIST when only the upper-half of an image is given.~\figref{fig:additional_mnist_2} shows MNIST completion results when 10, 100, and 1000 randomly selected pixels are given.

\clearpage
\subsection{Additional Image Completion Results on CelebA}
\begin{figure}[h!]
	\centering
	\includegraphics[width=\linewidth]{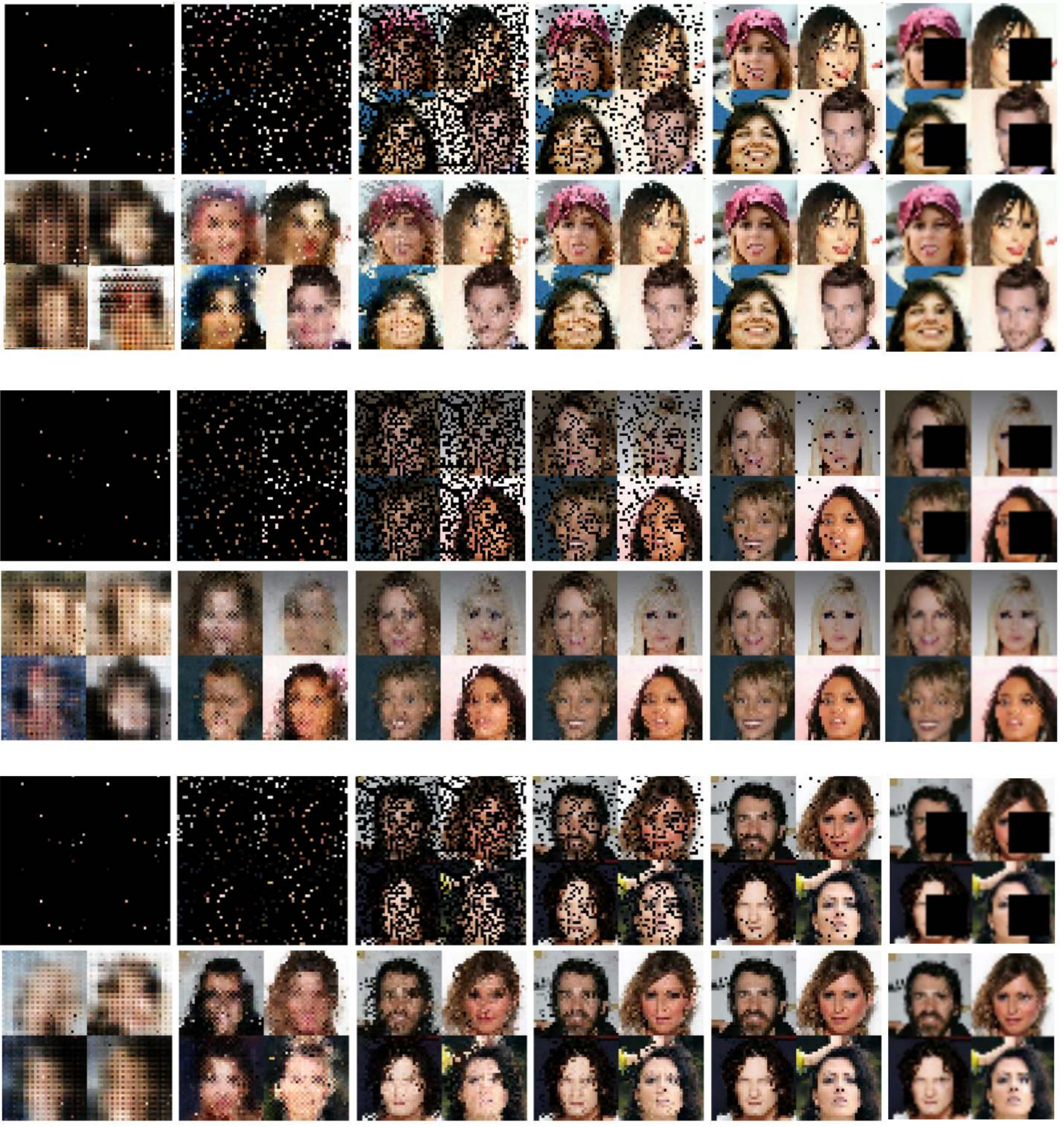}
	\caption{Additional image completion results on CelebA for both contiguous (right-most) and random pixels as context.}
  \label{fig:additional_celeba}  
\end{figure}

\figref{fig:additional_celeba} shows additional completion results on CelebA where $10, 100, 500, 800, 1000$ randomly selected pixels are given and when a $16$x$16$ square is removed from the original $32$x$32$ image.

\clearpage
\subsection{Additional Point-Cloud Generation Restuls}
\begin{figure}[h!]
\centering
\begin{subfigure}{1.\columnwidth}
  \centering
  \includegraphics[width=.8\linewidth]{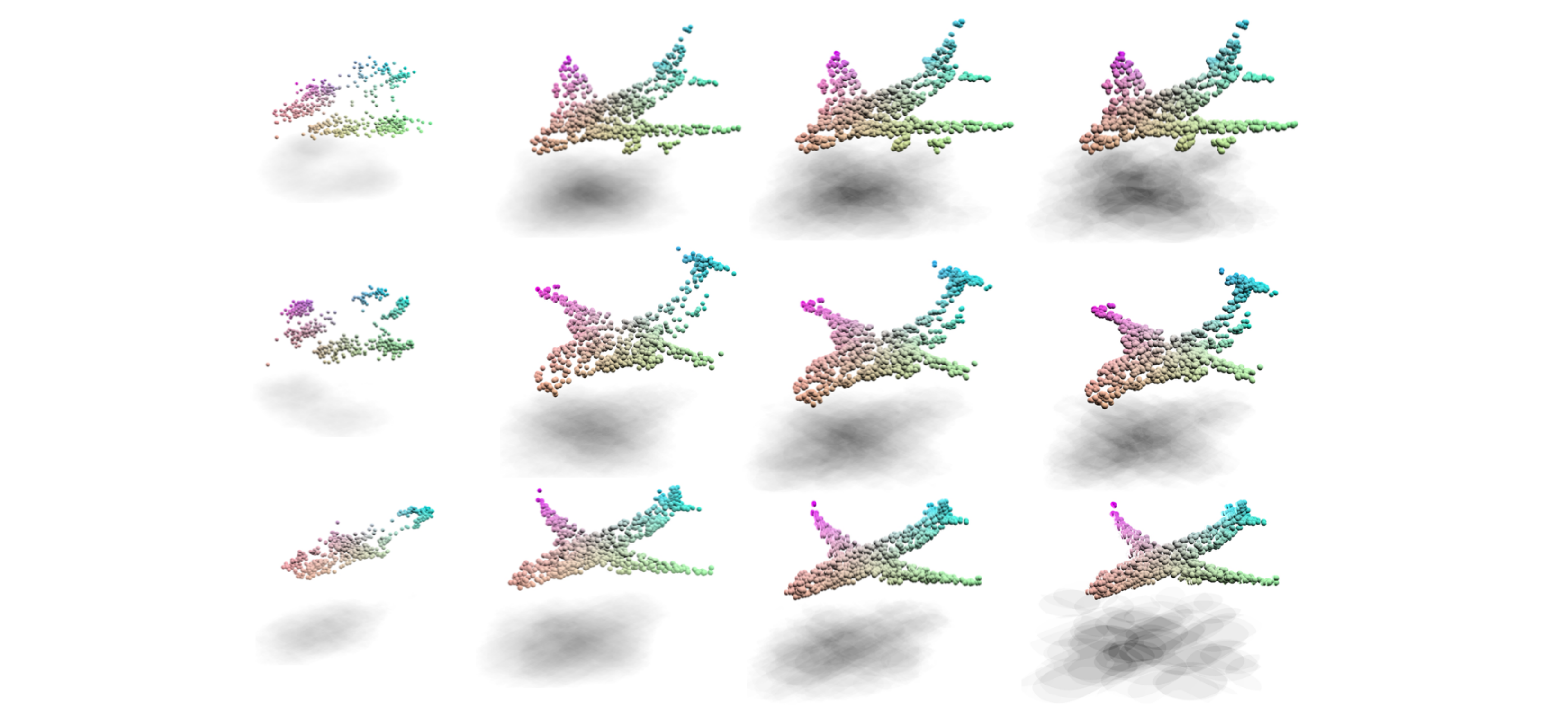}
\end{subfigure}
  \hfill
  
\begin{subfigure}{1.\columnwidth}
  \centering
  \includegraphics[width=.8\linewidth]{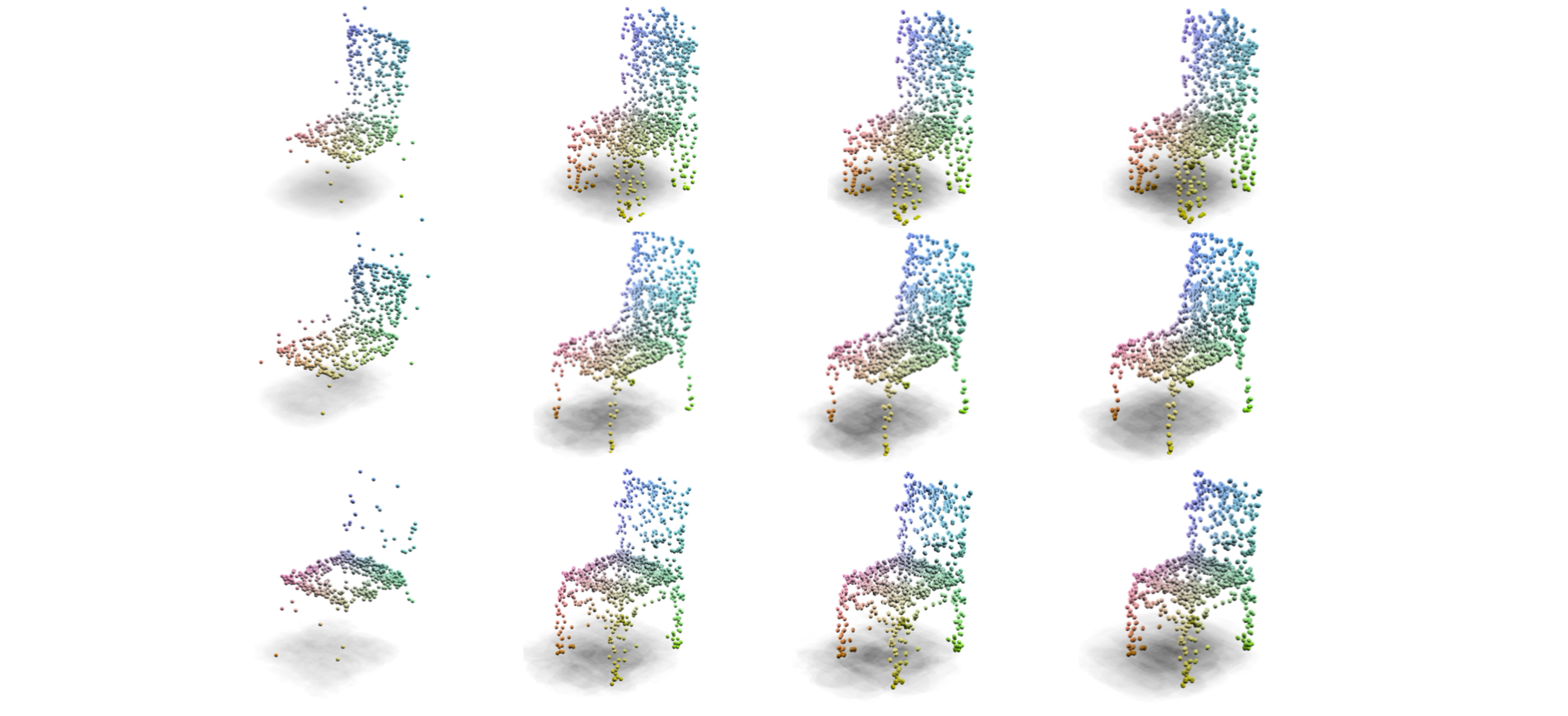}
\end{subfigure}
\begin{subfigure}{1.\columnwidth}
  \centering
  \includegraphics[width=.8\linewidth]{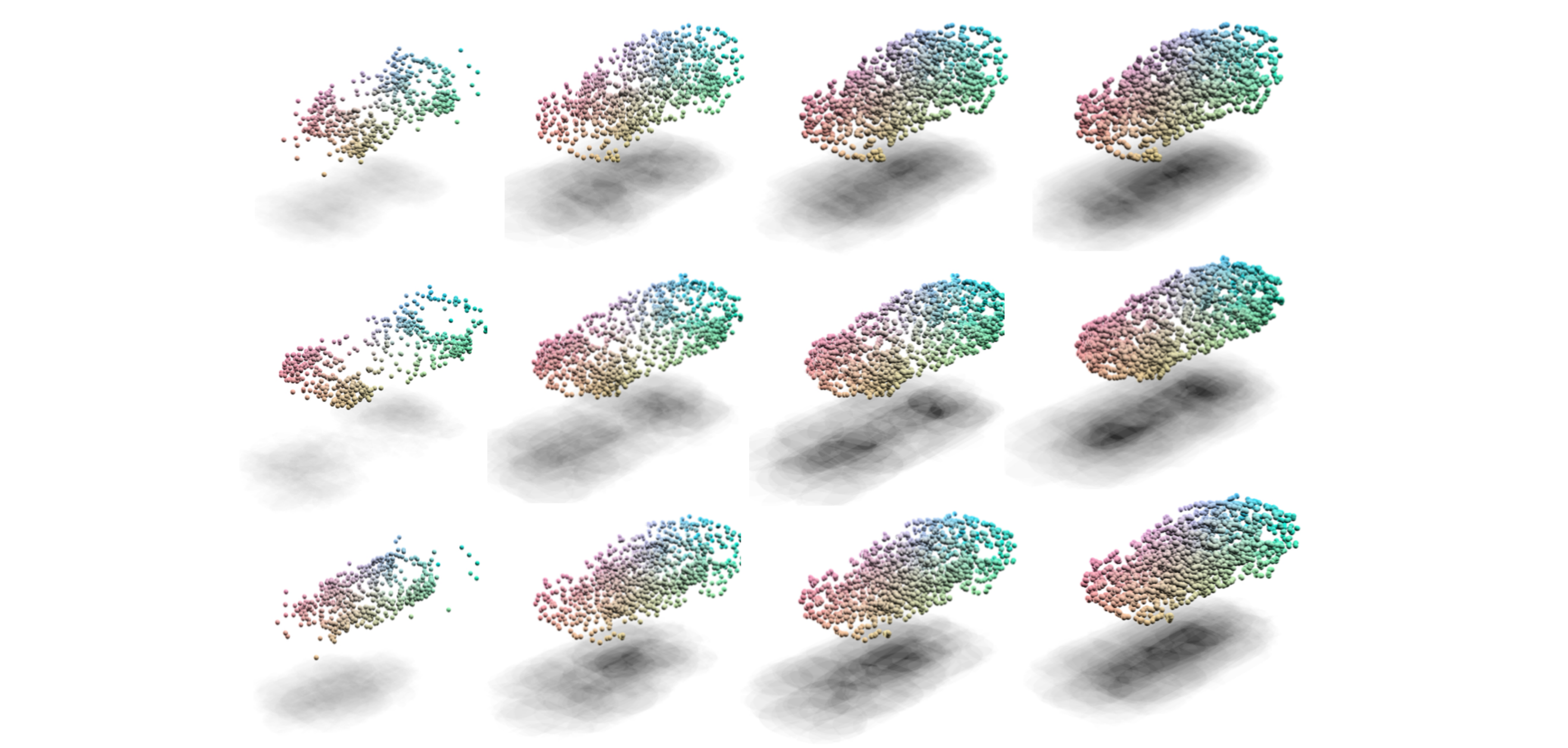}
\end{subfigure}

\caption{Additional examples of airplane, chair, and car point-cloud generation with 4 RNN blocks of block size 512.}
\label{fig:additional_pcgen} 
\end{figure}

\clearpage
\subsection{Additional Point-Cloud Denoising Restuls}
\begin{figure}[h!]
\centering
\begin{subfigure}{1.\columnwidth}
  \centering
  \includegraphics[width=.8\linewidth]{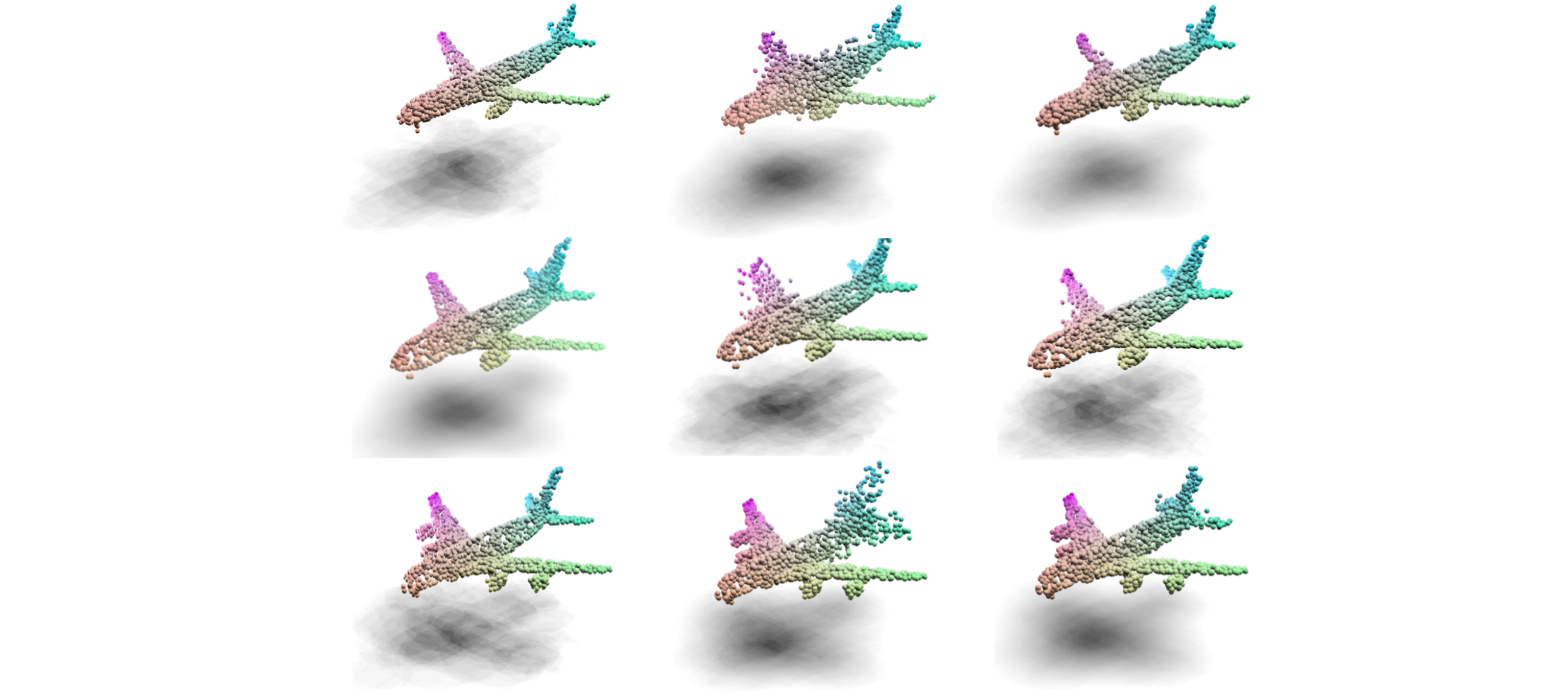}
\end{subfigure}
  \hfill
  
\begin{subfigure}{1.\columnwidth}
  \centering
  \includegraphics[width=.8\linewidth]{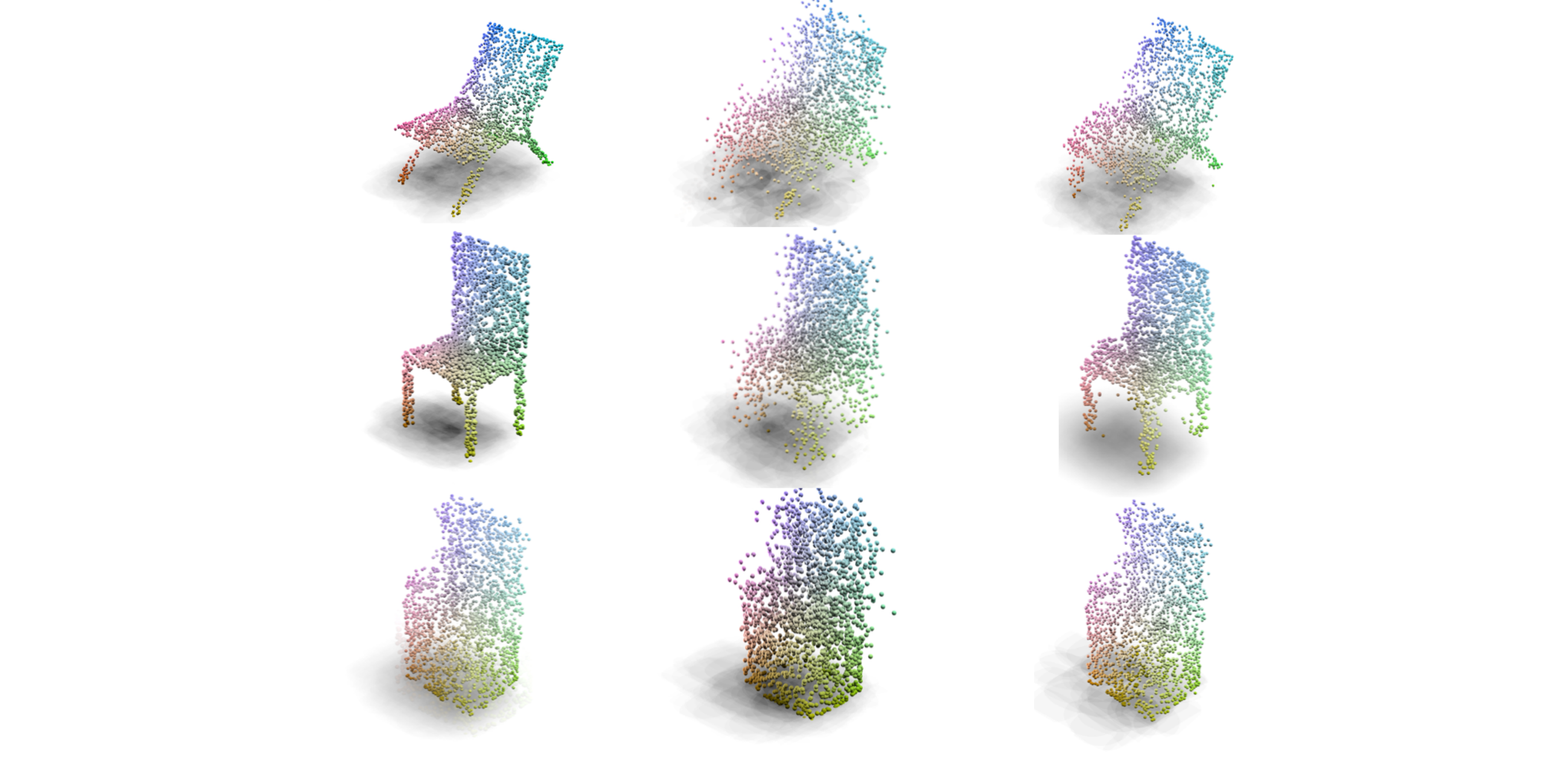}
\end{subfigure}
\begin{subfigure}{1.\columnwidth}
  \centering
  \includegraphics[width=.8\linewidth]{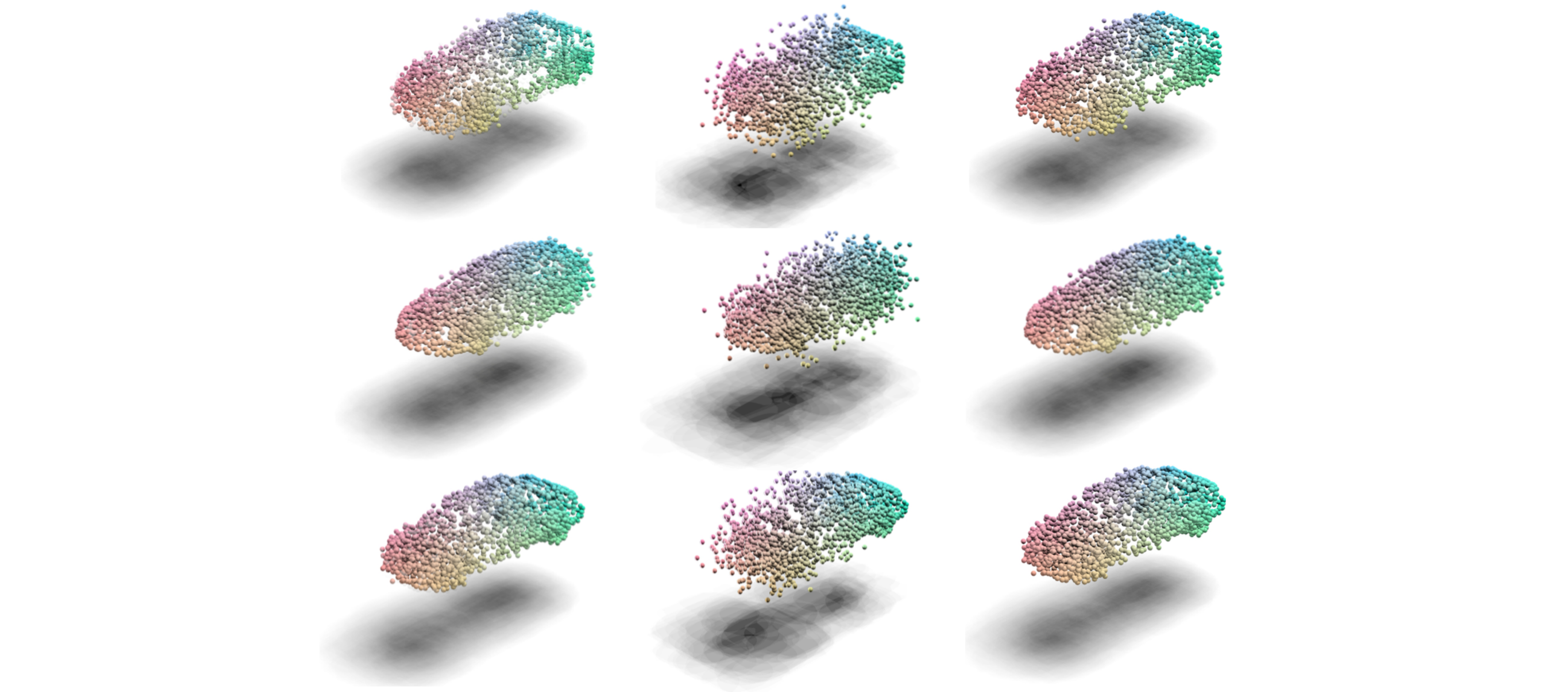}
\end{subfigure}

\caption{Additional examples of airplane, chair, and car point-cloud denoising. Left to right: original, perturbed, denoised point clouds.}
\label{fig:additional_pc_denoising}
\end{figure}

\end{appendix}
\end{document}